\documentclass[article,authoryear,jfv,table]{beg_32}             %  final version
\usepackage[hang]{footmisc}
\usepackage{xcolor}
\usepackage{soul}
\setcitestyle{round, comma, numbers, sort&compress, super}
\fancypagestyle{plain}{%
  \fancyhf{}
  \fancyhead[R]{\small {\it \jname}, x(x): \thepage--\pageref{LastPage} (\myyear\today)}
  \fancyfoot[R]{\small\bf\thepage }
  \fancyfoot[L]{\fottitle}
  }
\renewcommand{\dmyy}{17}
\renewcommand{\myyear}{2017}
\renewcommand{\today}{}
\newlength\mylen
\newcommand\myinput[1]{%
  \settowidth\mylen{\KwIn{}}%
  \setlength\hangindent{\mylen}%
  \hspace*{\mylen}#1\\}
\begin{document}

\volume{Volume x, Issue x, \myyear\today}
\title{Exploratory Lagrangian-Based Particle Tracing Using Deep Learning}
\titlehead{Exploratory Lagrangian-Based Particle Tracing Using Deep Learning}
\authorhead{Mengjiao Han, Sudhanshu Sane, \& Chris R. Johnson}
%For at least  authors with different addresses, use instead the following commands
% \corrauthor[1]{Mengjiao}
% \author[1]{S.B. Author, Jr.}
% \author[1]{Third Author}
% \corremail{mengjiao@sci.utah.edu}
% \corraddress{Business or Academic Affiliation 1, City, State, Zip Code}
% \address[1]{SCI Institute, University of Utah}
% \address[2]{Business or Academic Affiliation 2, City, Province, Zip Code, Country Business or Academic Affiliation 2, City, Province, Zip Code, Country}
% End information for at least  authors with different addresses
% For authors with the same post address,
\corrauthor{Mengjiao Han}
\corremail{mengjiao@sci.utah.edu}
\author{Sudhanshu Sane}
\author{Chris R. Johnson}
\address{University of Utah, SCI Institute, Salt Lake City, UT 84112, USA}
% End commands for all authors with the same address

%\dataO{mm/dd/yyyy}
%\dataO{}
%\dataF{mm/dd/yyyy}
%\dataF{}

\abstract{Time-varying vector fields produced by computational fluid dynamics simulations are often prohibitively large and pose challenges for accurate interactive analysis and exploration.
To address these challenges, reduced Lagrangian representations have been increasingly researched as a means to improve scientific time-varying vector field exploration capabilities.
This paper presents a novel deep neural network-based particle tracing method to explore time-varying vector fields represented by Lagrangian flow maps.
In our workflow, in situ processing is first utilized to extract Lagrangian flow maps, and deep neural networks then use the extracted data to learn flow field behavior.
Using a trained model to predict new particle trajectories offers a fixed-small memory footprint and fast inference.
To demonstrate and evaluate the proposed method, we perform an in-depth study of performance using a well-known analytical data set, the Double Gyre.
Our study considers two flow map extraction strategies as well as the impact of the number of training samples and integration durations on efficacy, evaluates multiple sampling options for training and testing, and informs hyperparameter settings. 
Overall, we find our method requires a fixed-memory footprint of 10.5 MB to encode a Lagrangian representation of a time-varying vector field while maintaining accuracy.
For post hoc analysis, loading the trained model costs only two seconds, significantly reducing the burden of I/O when reading data for visualization.
Moreover, our parallel implementation can infer one hundred locations for each of two thousand new pathlines across the entire temporal resolution in 1.3 seconds using one NVIDIA Titan RTX GPU.
}


\maketitle

\section{Introduction}
Numerical flow visualization plays a critical role in enabling scientists to understand fluid phenomena and improve computational fluid dynamics models.
Although simulations typically produce time-varying vector fields, analysis and visualization are often limited to single time slices due to I/O constraints and memory requirements.
Performing accurate time-varying flow visualization using traditional methods requires a high temporal resolution of the vector field data.
A potential solution to perform accurate time-varying flow visualization is to consider a Lagrangian representation of the vector field.
%
%Although Lagrangian representations have extensively been used by the ocean modeling community~\cite{VANSEBILLE201849}, in recent years, several research works have studied the use of reduced Lagrangian representations to enable time-varying flow visualization~\cite{agranovsky2014improved,bujack2015lagrangian,chandler2015interpolation,chandler2016analysis,sane2018revisiting,sane2019interpolation}.
%
Lagrangian representations have been demonstrated to offer strong accuracy-storage propositions compared to traditional techniques (\citet{agranovsky2014improved,sane2021investigating}).
The approach involves two phases: in situ and post hoc.
Lagrangian representations are extracted from computational simulations using in situ processing and explored during post hoc analysis.
In this paper, we study the use of deep learning methods to perform post hoc exploration of time-varying vector fields using reduced Lagrangian representations computed in situ as training data. 

In recent years, the scientific visualization community has seen an increased adoption of deep learning (\cite{leventhal2019pave,weiss2019volumetric,berger2018generative,hong2019dnn,he2019insitunet,han2019tsr,han2020v2v,engel2020deep}), including multiple research projects that consider vector field data (\cite{han2018flownet,han2019flow,Jakob2020,sahoo2021integration,guo2020ssr,kim2019deep,liu2019cnn}).
With respect to exploratory Lagrangian-based particle advection schemes, the use of deep learning has not previously been studied to the best of our knowledge.
Prior strategies have relied on constructing search structures over the data to identify sets of precomputed particle trajectories that can be interpolated across intervals of time.
Search structures such as k-d trees and Delaunay triangulations can be computationally expensive to compute for each interval and memory intensive for large data sets (\cite{hlawatsch2011hierarchical, chandler2015interpolation, sane2019interpolation}).
Our study shows that, by leveraging deep learning, we can limit the memory footprint of the extracted data.
Importantly, once the model is trained, it provides quick inference of new particle trajectories during post hoc analysis and exploration.

Overall, we contribute the first deep neural network-based method to encode Lagrangian flow maps and enable exploratory particle tracing in time-varying flow fields.
Our study demonstrates the performance of the method across varying hyperparameter settings as well as multiple Lagrangian representation configurations. 
Our trained model requires a fixed-memory footprint of 10.5 MB, potentially offering a potentially significant data reduction for high-resolution flow maps and alleviating I/O costs during exploration. 
Further, the trained model can infer new trajectories accurately and at rates supporting interactive exploration.
Lastly, we consider a widely studied analytical data set, the Double Gyre, as well as, a second vector field targeted to machine learning applications to demonstrate our approach.
%

%\mengjiao{Revised main contributions}
%Our main contributions are:
%\begin{enumerate}[itemsep=2mm, parsep=1.5pt]
%    \item Our work is the first to propose and evaluate leveraging deep neural networks to encode Lagrangian flow maps for exploratory particle tracing in time-varying flow fields. 
%    \item Our trained model can reduce the memory footprint of high-resolution flow maps from arbitrary data sets to 10.5 MB while achieving accurate inference. 
%    \item Using the trained model costs only two seconds for loading, which significantly reduces the overhead of I/O. 
%    % \item We provide a visualization tool integrating the OSPRay renderer and our trained model to produce high-fidelity visualization of reconstructed trajectories.   
%\end{enumerate}

\section{Related Work}
This section provides background on Lagrangian analysis, the use of reduced Lagrangian representations, and the use of machine learning for flow visualization tasks.

\subsection{Lagragian Analysis}
Lagrangian analysis is a powerful tool, widely adopted by the ocean modeling community (\cite{VANSEBILLE201849}), to explore time-varying vector fields generated by simulations.
In response to growing data set sizes, reduced Lagrangian representations have been increasingly researched as a solution to enable time-varying vector field exploration across various application domains.
Reduced Lagrangian representations are computed using in situ processing and explored during post hoc analysis. 
By utilizing in situ processing, Lagrangian representations are computed using the complete spatial and temporal resolution of the simulation data.
Studies have demonstrated reduced Lagrangian representations offer strong accuracy-storage propositions for exploration in temporally sparse settings (\cite{agranovsky2014improved, rapp2019void, sane2021investigating}) as well as directly support feature extraction (\cite{froyland2015rough,schlueter2017coherent,hadjighasem2017critical,froyland2018robust,Jakob2020}).
Additionally, previous research has demonstrated the traditional Eulerian paradigm performs poorly in under-resolved temporal settings (\cite{costa2004lagrangian,Qin2014,agranovsky2014improved,sane2018revisiting,rockwood2019practical,sane2021investigating}).

In the Lagrangian specification of a time-varying vector field, information is encoded using particle trajectories.
Thus, the Lagrangian representation consists of a collection of particle trajectories spanning the spatial domain and can be defined as a flow map.
The flow map $F_{t_0}^{t}(x_0):\mathbb R^d \times \mathbb R \times \mathbb R  \to \mathbb R^d$ describes where a massless particle starting at position $x_0\in \mathbb R^d$ and time $t_0\in \mathbb R$ moves in the time interval $[t_0,t]\subset \mathbb R$ (\cite{garth2007efficient}).

Research related to reduced Lagrangian representations that enable time-varying vector fields has advanced along multiple axes.
These include in situ sampling techniques (\cite{agranovsky2014improved, rapp2019void, sane2019interpolation, sane2021scalable}), post hoc reconstruction strategies (\cite{hlawatsch2011hierarchical, agranovsky2015multi, bujack2015lagrangian, chandler2015interpolation}), theoretical and empirical error analysis (\cite{chandler2016analysis, hummel2016error, sane2018revisiting}), feature extraction (\cite{froyland2015rough,schlueter2017coherent, hadjighasem2017critical,froyland2018robust,Jakob2020}), and application to various domains (\cite{envirvis.20171099, siegfried2019tropical,sane2021investigating}).
In this paper, we study the use of deep learning to perform post hoc reconstruction. %when considering Lagrangian representations extracted using two in situ sampling techniques.
Specifically, we propose and evaluate the use of multi-layer perceptrons (MLPs) to learn the time-varying vector field behavior from previously computed particle trajectories.
With deep learning, a model can be trained once and then be interactively queried at the time of exploration without the significant memory requirements of prior approaches.
Our study focuses on the impact of various hyperparameters and extraction configurations on the efficacy of post hoc reconstruction as well as the overall computational cost.
%
% In fluid dynamic, there are two frames of references for 
% representing flow behaviors: Eulerian and Lagragian. 
% In Eularian frame of reference, velocity field is saved directly at a fixed time interval. 
% In contrast, Lagragian frame of reference uses flow map F for encoding flow behaviors. 
% A flow map saves massless particels' start locations at a start time, and the end locations over an interval of time. 
% The end locations are calculated by advected through velocity fields. 
% The last decade has seen a growing trend towards using Lagragian-Based method. 
% Compared to Eulerian representation, Lagragian representation has been demonstrated to be up to 
% 10X more accurate when consumes the same storage and can save up to 64X storage space when provide 
% similar accuracy under sparse time settings. 

\subsection{Flow Visualization Using Machine Learning}
In recent years, machine learning techniques have been increasingly researched by the fluid dynamics community (\cite{brunton2020machine}).
Similarly, with respect to scientific visualization, specifically, flow visualization, the use of machine learning to perform several tasks has increased.
For example, it has been widely used to detect flow field features such as eddies and vortices (\cite{lguensat2018eddynet, yi2018cnn, strofer2018data, bai2019streampath, duo2019oceanic, liu2019cnn,deng2019cnn, wang2021rapid}).
\cite{kim2019robust} utilized the convolutional neural networks (CNNs) to extract a robust frame of reference for unsteady two-dimensional (2D) vector fields.
\cite{hong2018access} used the long short-term memory (LSTM) to improve data access patterns for improved computational performance during distributed memory particle advection.
\cite{li2015extracting} employed the support vector machine (SVM) to segment streamlines based on user-identified features.
For the widely studied task of selecting a representative set of particle trajectories (\cite{sane2020survey}), recent state-of-the-art techniques by~\cite{han2018flownet} and~\cite{lee2021deep} have used deep-learning-based clustering approaches.
Further, modern techniques to reconstruct steady state vector fields using a set of streamlines employ machine learning (\cite{han2019flow,sahoo2021integration}).

\cite{Jakob2020} upsampled 2D finite-time lyapunov exponent (FTLE) scalar fields derived from Lagrangian flow maps using an efficient subpixel convolutional neural network (ESPCN) by \cite{shi2016real} and SRCNN by~\cite{dong2015image}.
In our study, we use the Lagrangian representations of 2D time-varying vector fields as data to train neural networks built with MLPs.
We then infer new particle trajectories from the model to support the exploration use case.
Our study shows that the application of deep learning to particle tracing can offer the significant benefits of reduced memory requirement and accurate trajectory inference.
%can significantly reduce the memory footprint of high-resolution flow maps and can reconstruct accurate results.

\begin{figure}[!htb]
    \subfigure[The workflow of our proposed approach. The Lagrangian flow maps are calculated using in situ processing and saved to the database. 
    The network is trained using the particle start locations and the corresponding end locations at various file cycles. Once the model is fully trained, new particle trajectories can be inferred from the model.\label{fig:6a}]
    {\includegraphics[width=\textwidth]{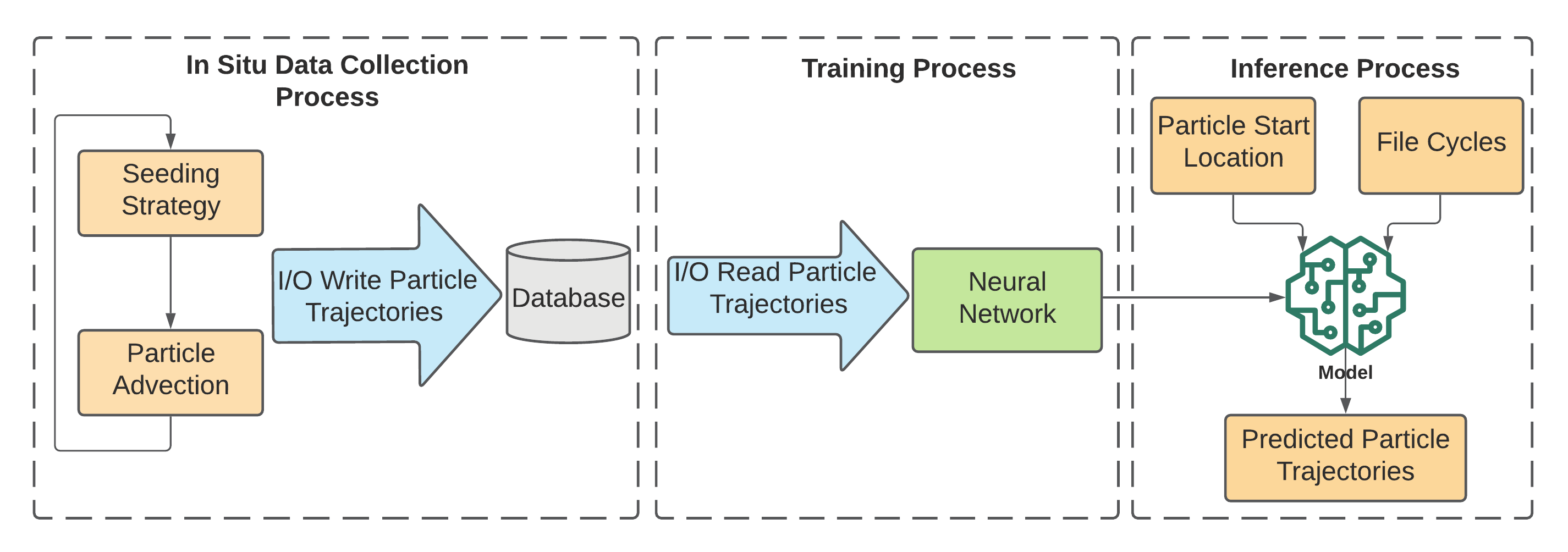}}
    \quad
    \subfigure[\label{fig:6b}The architecture of our neural network built with multi-layer perceptrons (MLP). The network takes the particle start location and the file cycles as input, and outputs the particle end locations. ]
    {\includegraphics[width=\textwidth]{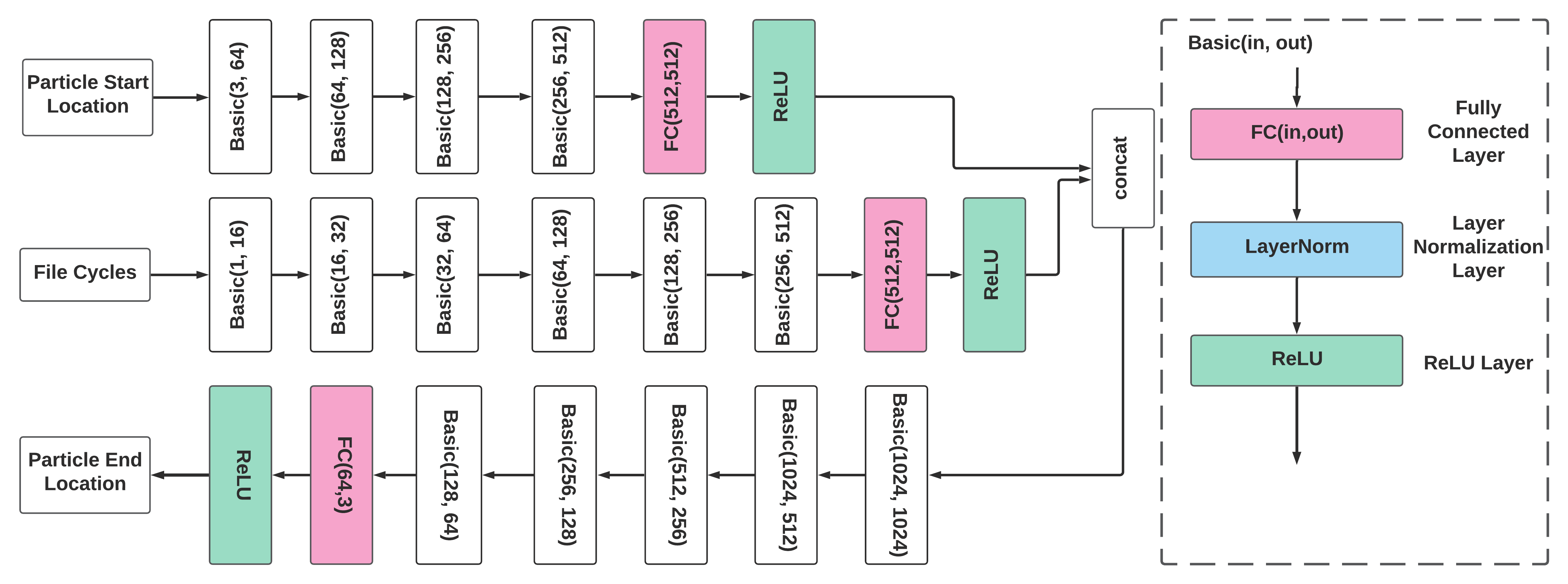}}
    \caption{\label{fig:overview} Unlike prior two-phase Lagrangian analysis workflows, after extracting Lagrangain representations using in situ processing, a preprocessing phase involving neural network training is introduced prior to post hoc analysis. Figure~\ref{fig:6a} shows the high-level workflow of our proposed approach and Figure~\ref{fig:6b} shows the details of the neural network architecture.}
\end{figure}

\section{Lagrangian Analysis using Deep Learning}

We designed our network to learn the flow behavior encoded by the Lagrangian representation of the time-varying vector field. 
% Given a particle in the domain, the trained model can answer where the particle ends 
% after a specific . 
Figure \ref{fig:6a} shows the workflow of in situ training data generation process, network training process, and the post hoc inference process.
In the in situ extraction phase, Lagrangian flow maps are computed by advecting particles using 
%Euler's method and 
the full spatial and temporal resolution of the time-varying vector field.
%from the simulation equations. 
We considered two approaches to extract flow maps,
\begin{itemize}
\item $Lagrangian_{long}$:  extract a single flow map consisting of long particle trajectories with a uniform temporal sampling of each integral curve. 
%extract a single flow map with saving multiple intermediate locations along the trajectories; 
\item $Lagrangian_{short}$: extract multiple short flow maps with each flow map consisting of a set of seed locations and a set of end locations for each seed, where each end location in a set corresponds to the displacement from the seed location over non-overlapping intervals of time.
%
%multiple end locations of particle trajectories integrated over non-overlapping intervals of time. 
\end{itemize}
%extract multiple flow maps with terminating, saving end locations, and resetting start seeds. 
%
In our paper, we follow the notation used by~\cite{agranovsky2014improved}. We refer to the cycles where the end location is saved out as \textit{file cycles}. 

To begin the post hoc analysis phase, the network fetches flow maps from the database, pre-processes them, and loads data as training samples (Section \ref{sec:training_data}). 
The network architecture is built with MLP that are a series of fully connected layers (Section \ref{sec:network}). 
The loss function is set to the L1 loss, which is calculated as the error between the target end location and the predicted end location.
% In addition, we minimize the angle loss, which is calculated as the cosine similarity 
% %
% between the vectors of the target and predicted end location~(Section \ref{sec:loss}).
%between the vector of target end location and predicted end location 
%to the end location at the previous file cycle (Section \ref{sec:loss}). 
%
During the training process, the model takes two parameters, particle start locations and queried file cycles as inputs, and outputs the corresponding end locations.
Weights of the model are updated by backpropagation of the loss to find the optimized weights~(Section~\ref{sec:training_process}). 
Finally, new trajectories can be infered from the trained model~(Section~\ref{sec:inference}).

\subsection{Training Data Generation}
\label{sec:training_data}
We stored extracted Lagrangian flow maps in the form of training data for the model.
%
%The training data generation process is the same as the process of Lagrangian-Based flow maps extraction. 
%
%We compute Lagrangian flow maps by advecting particles using Euler's method and the complete temporal resolution of the vector field.
%from the simulation code. 
We considered two strategies to sample the time-varying vector field.
The first strategy, $Lagrangian_{long}$, involves computing long trajectories with uniform sampling along the curve.
Reconstruction of new trajectories using long precomputed trajectories is more accurate when the propagation of error is eliminated after every interpolation step (\cite{hummel2016error,sane2019interpolation}).
However, the quality of domain coverage may be reduced as the integration time increases due to divergence in the flow field (\cite{chandler2016analysis}).
The second strategy, $Lagrangian_{short}$, involves computing sets of short trajectories with only the start and end location after non-overlapping intervals of time stored.
Although such an approach offers improved domain coverage (\cite{agranovsky2014improved}), the particle trajectory reconstruction may be less accurate due to error propagation (\cite{bujack2015lagrangian}). 

%(1) extract a single flow map consisting of long particle trajectories with uniform sampling of the integral curve.
%(2) extract multiple short flow maps with each flow map consisting of a set of seed locations and the corresponding multiple end locations of particle trajectories integrated over non-overlapping intervals of time.

For both approaches, the first step is placing sample seeds in the domain.
In this paper, we denote the number of seeds by $N$. 
To understand the impact of the seed placement strategy on the model inference performance, we studied three strategies: (1) seeding along a uniform grid ($uniform$), (2) seeding using a pseudorandom number sequence ($random$), and (3) seeding using a Sobol quasirandom sequence ($sobol$). 
Specifically, we considered reconstruction accuracy near features of interest and boundaries.
Although placing uniform seeds can provide good domain coverage and fast interpolation during post hoc analysis, it does not optimize information per byte stored.
Thus, in many practical cases, the Lagrangian representation can be unstructured and would typically incur a higher interpolation cost during post hoc analysis.
By considering $random$ and $sobol$ seeding, we were able to demonstrate the fast inference of new trajectories from unstructured Lagrangian flow maps.
%
%respect to optimizing for information per byte and not representative of non-uniform sampling techniques. 
%
%Pseudorandom number sequence and Sobol quasirandom sequence can apply to structured and unstructured grids and compare to 
%pseudorandom number sequence, Sobol quasirandom sequence tend to sample more uniformly~\cite{Burkardt_Chisari_Fox}. 
%
We compare these three seeding choices in Section~\ref{sec:seeding_strategy}.
% In order to sample the entire domain using an unstructured set of points, we use the sobol quasirandom sequence, 
% which tend to sample more uniformly than the pseudorandom number sequence~\cite{Burkardt_Chisari_Fox}.
%\sudhanshu{is there a reason we could not use uniform samples? also I did not know this seed strategy allows learning of maximal flow? Is there a citation for that?}
%
%The benefit of using this seed strategy is obvious that the model is able to learn maximal flow behaviors. 
%
% However, the drawback of this seed placement strategy is it fails to place seeds exactly on the boundary, which can results in inaccurate extrapolation near the boundary of the domain.
%to extrapolate is not able to 
%capture the flow behaviors of the boundary. 
%Further work is required to improve this seed strategy and for the reconstruction of new trajectories seeded on the boundary. 
%

After seeds are placed, particle trajectories are computed by displacing particles from time $t$ to $t+\delta$, where $\delta$ indicates an advancement by one simulation time step.
Following the notation in~\cite{agranovsky2014improved}, we refer to one simulation advancement as a \textit{cycle}, 
the cycle on which the simulation saves data as a \textit{file cycle}, and the number of cycles between file cycles as the \textit{interval} 
in the following sections. 
Given a total temporal duration $T$, the total number of file cycles 
$n$ can be calculated by 
\begin{equation}
    \label{eq:n}
% \begin{align*}
    \begin{aligned}
    n &= floor(T / (\delta * C)) 
    % \text{where}~C~\text{represents file cycles \textit{interval}}
    \end{aligned}
% \end{align*}
\end{equation}
\sloppy where $C$ represents the file cycle \textit{interval}. Thus, the list of file cycles is $C_{0:n-1} = [C,\, 2C, \,3C, \,..., \,nC]$.   
%When applying the first approach of generating flow maps, 
To generate $Lagrangian_{long}$ flow maps, seeds are placed once at the beginning at time 
$t_0 = 0$ and traced until $T$, i.e., the entire temporal duration. 
Intermediate locations are recorded along each trajectory at every file cycle. 
To generate $Lagrangian_{short}$ flow maps, particle tracing starts at time $t_0 = 0$ and terminates at time $t_1 = t_0 + \delta * C$.
Then, the location at $t_1$ is saved, and seeds are reset for the tracing until the next file cycle. 
This process is repeated until the last file cycle. 

% In the paper, We denote the interval of file cycles by $C$, 
% the temporal duration by $T$ and one cycle duration by $\delta$.
% The total number of writing cycles $n$ can be calculated by \[n=floor(T / (\delta * C))\]
% The list of writing cycles is $C_{0:n-1} = [C,\, 2C, \,3C, \,..., \,nC]$. 

The training data sets are saved in the NPY file format for efficient loading in Python. 
We created a three-dimentional (3D) array, with dimensions of $[n+1, N, 3]$, for saving start seed locations and corresponding end locations at various file cycles.
When loading the data sets, the data are organized into training samples, as shown in Equation \ref{eqn:inputs}.
One training sample contains start location $start_i$~(where $i = 0, 1, ... N-1$), 
the queried file cycle $C_{j}$~(where $j = 0, 1, ... n-1$), and the 
target end location at the queried file cycle $target_{i, j}$~(where $i = 0, 1, ... N-1$ and $j = 0, 1, ... n-1$).
%and the end location at the previous file cycle($prev_{i,j-1}$).
%
%\sudhanshu{$pred_{i,j-1}$ --- this seems like it shouldn't be part of a training sample..}
The start location and the queried file cycle are inputs to the network.
The target end locations are used for calculating the loss (Equation \ref{eq:l1}).
In addition to training data, we generated validation data by 
using $0.1 * N$ seeds~(10\% of training samples) and following the same process.

\begin{equation}
    \label{eqn:inputs}
    \begin{aligned}
    Inputs = & \{\{start_0, \,C_0, \,target_{0, C_0}\}, \\
             & \{start_0, \,C_1, \,target_{0, C_1}\}, ..., \\
             & \{start_0, \,C_{n-1}, \,target_{0, C_{n-1}}\}, ..., \\
             & \{start_{N-1}, \,C_{n-1}, \,target_{N-1, C_{n-1}},\}\} 
    \end{aligned}
\end{equation}

% Where $start_i$ is ith particle start location at cycle 0 and $C_{j}$ is the queried writing cycle.
% $end_{i, C_{j}}$ is the end location of ith start seed at queried writing cycle $C_{j}$ that is used for calculating 
% L1 Loss (Section \ref{sec:loss}) and $pre_{i,C_{j-1}}$ is the end location of ith start seed 
% at the previous writing cycle that is used for calculating angle loss (Section \ref{sec:loss}). 

\subsection{Network Architecture}
\label{sec:network}
The network architecture, shown in Figure \ref{fig:6b}, consists of a latent encoder $E$ and a latent decoder $D$.
The latent encoder $E$ and decoder $D$ are built with MLP, a series of fully connected layers.
The latent encoder $E$ takes a particle's start location $start$ and a queried file cycle $C_{j}$ as inputs.
These two parameters are separately fed into two sequences of fully connected layers of size (64, 128, 256, 512) and (16, 32, 64, 128, 256, 512).
The two outputs are then concatenated together as a latent vector. 
Next, the latent decoder $D$ that is also a series of fully connected layers of size (512, 256, 128, 64) is followed by the latent vector being mapped to predicted end location $pred$ at the queried file cycle.
We added layer normalization (\cite{ba2016layer}) after each fully connected layer except output layers to stabilize the training process.
Moreover, we used the rectified linear unit (ReLU) (\cite{nair2010rectified}) as the activation function for each output from the fully connected layer. 

% \subsection{Loss Function}
% \label{sec:loss}
% We applied L1 loss as loss functions in our method. 
% %
% \textbf{L1 loss} calculates the mean absolute error between target and predicted end locations by the network (Equation~\ref{eq:l1}).
% 
% \textbf{Angle loss} (Equation~\ref{eq:angle}) calculates the cosine similarity between the vectors from the start location to the target and predicted end locations. 
%
% Using the angle loss guides our network to predict end locations with a similar direction to the ground truth.
%

% \begin{equation}
%     \label{eq:l1}
%     L1\_Loss = |end - \hat{end}|
% \end{equation}
% \begin{equation}
%     \label{eq:angle}
%     \begin{aligned}
%         Angle\_Loss = & 1 - cosine\_similarity(|end_{i, C_j} - start_{i, C_j}|, \\
%                       & |\hat{end_{i, C_j}} - start_{i, C_j}|) 
%     \end{aligned}
% \end{equation}
% where $i$ is the seed index and $j$ is the file cycle index.

% The total loss $L$ combines the L1 loss and the angle loss,
%
% \begin{equation}
% \[ L = L1\_loss \;+ \;Angle\_loss\]
% \end{equation}

\subsection{Training Process}
\label{sec:training_process}
\vspace{-1em}
\begin{algorithm}[!ht]
    \KwIn{Data set shown in Equation \ref{eqn:inputs}}
    \myinput{Initial weights of the network $w$}
    \KwOut{Optimized weights $w$}
    
    Load training data set 

    \For{each epoch} 
    {
        \For{each batch of training samples}
        {   
            model.train()\\
            $pred = model(start, queried\_file\_cycle)$\\
            $loss = L1\_Loss(pred, target)$ \\
            Backpropagation and update weight $w$ 
        }
        \For{each batch of validation samples}
        {   
            model.eval()\\
            $pred = model(start, queried\_file\_cycle)$\\
            $loss = L1\_Loss(pred, target)$
        }
        call learning rate scheduler adjust the learning rate if needed
    }
    
    \caption{{\bf Training Process} \label{Algorithm:training}}
    
\end{algorithm}

We implemented our neural network using Pytorch (\cite{NEURIPS2019_9015}). 
The training process, shown in Algorithm \ref{Algorithm:training}, aims to find the 
optimized weights $w$ of the network. The weights are initialized by Pytorch. 
We created a custom Pytorch \textit{Dataset} class to load and store all training samples.   
We then loaded the Pytorch \textit{Dataset} object into a Pytorch \textit{DataLoader} for 
iterating through the training samples. At the beginning of each epoch, the training samples are shuffled and 
split into batches. Given a batch of training samples, the forward process computes the output 
following the network architecture and computes the loss as defined by the loss function. 
The backpropagation process is done automatically using Pytorch by calling \textit{loss.backward()}, and the weights are updated by the optimizer. 
For our experiments, we trained the network for 100 epochs using the Adam optimizer (\cite{kingma2014adam})
with the hyperparameters of $\beta_1 = 0.9$, $\beta_2 = 0.999$, and $\varepsilon = 1e-6$. 
Further, in our training process, we set the initial learning rate to $10^{-5}$ and 
used a learning rate scheduler (\cite{ReduceLROnPlateau}), 
provided by Pytorch to reduce the current learning rate by a factor of 2 
if the validation loss had not decreased for five epochs. 
We applied L1 loss as loss functions in our method. 
\textbf{L1 loss} calculates the mean absolute error between target and predicted end locations by the network (Equation~\ref{eq:l1}).

\begin{equation}
    \label{eq:l1}
    L1\_Loss = |target - pred|
\end{equation}

\subsection{Inference Process}
\label{sec:inference}
Besides varying generation processes for $Lagrangian_{long}$ and $Lagrangian_{short}$, the inference process when using the model trained by data from these two approaches also varies.
When using $Lagrangian_{long}$, interpolations are performed by always considering the new seed start location at $t_0=0$.
The end location inferred by the model results from the provided start location and the queried file cycle.
In contrast, when using $Lagrangian_{short}$, new particle trajectories are ``stitched'' together by advancing the new seed across intervals.
Here, the inference is performed by considering the location of the seed particle at the previous file cycle and the target file cycle.
Since every inference except the first uses previously inferred results, errors might propagate along new trajectories when using $Lagrangian_{short}$ (\cite{hummel2016error, sane2019interpolation}).
We refer to the absolute error introduced by the model for any single inference as \textit{local error} and to the error accumulated along particle trajectories that are ``stitched'' together as \textit{global error}.
Similar to other Lagrangian-based advection schemes, our inference process currently is limited to interpolating the locations along a particle trajectory at file cycles, and in the case of $Lagrangian_{long}$, it is limited to particles starting at $t_0=0$.
%We refer to \textit{with error propagation} as updating every inference using previous inferred results and \textit{without error propagation}
%as the absolute error introduced by the model in following results.

To measure the accuracy of new particle trajectories inferred by the model, we calculated the average of aggregated Euclidean distance between the target ground truth and the model predicted along each trajectory (Equation \ref{eq:measure}). 
\begin{equation}
    \label{eq:measure} 
    error_i = \frac{1}{n}\sum_{j=0}^{n} L_{2}(target_j - pred_j) 
\end{equation}
where $i$ represents the index of the new seed and $n$ is the number of end locations (file cycles) along the trajectories (Equation \ref{eq:n}).

% To measure the accuracy of new particle trajectories inferred by the model, we used a robust and accurate metric called the adaptive edit distance on real sequences (AEDR) proposed by ~\cite{ren2020uncertainty} to measure pathline similarity.
% %
% The metric uses the L1 norm divided by a threshold distance to quantify the local error of each interpolated location (Equation \ref{eq:aedr}), accumulates error along the trajectory, and produces an average across all the interpolated locations.
% %
% \begin{equation}
%     \label{eq:aedr}
%     AEDR(end, \hat{end}) = 
%     \begin{cases}
%         L_{1}(end, \hat{end}) / \zeta,& \text{if } L_{1}(end, \hat{end}) < \zeta\\
%         \zeta,              & \text{otherwise}
%     \end{cases}
% \end{equation}
% where $\zeta$ is the value of distance threshold.

% The AEDR metric is based on edit distance on real sequences (EDR) and has been demonstrated less sensitive to the outliers and can be applied to pathlines with different length.
% %
% The use of a threshold distance and maximum error at any particular sample results in an AEDR error value between 0 and 1. 
% %
% A value close to 0 indicates the particle trajectories are similar, whereas a value close to 1 indicates the particle trajectories are dissimilar.
% %
% In our experiments, we used $\zeta = 1$ to simplify the problem. 

\section{Results}
In this section, we first describe the data set used for our experiments~(Section~\ref{sec:data_sets}). 
Next, we present an evaluation of sampling strategies and hyperparameters (learning rate, batch size) used during training data generation~(Section~\ref{sec:hyper-parameters}), followed by a report of the performance of our proposed network for training and inferences~(Section \ref{sec:network_training}). 
Finally, to evaluate the accuracy of the model across Lagrangian flow map extraction parameter settings, we quantitatively and qualitatively evaluate the impact of varying the number of seeds~(Section~\ref{sec:num_seeds}) and file cycle intervals~(Section~\ref{sec:writing_cycles}).
%
% Finally, we evaluate the performance of our method applied to an ensemble data set~(Section \ref{}).
%

\subsection{Data Set}
\label{sec:data_sets}
We conducted our study by considering a standard benchmark data set frequently used to study fluid dynamics and, in particular, flow visualization tools and techniques: the 2D unsteady Double Gyre~\cite{Shadden05}. 
The model of the unsteady Double Gyre flow field is widely studied for the computation of hyperbolic Lagrangian coherent structures (LCS) in flow data. 
%
%The process of generating data is described in detail in Section \ref{sec:training_data}.
%
For all the training data generated, we considered a total temporal duration of $[0, 10]$ with $\delta = 0.01$.
The Double Gyre flow field is defined by equation~\ref{eqn:Double_Gyre} within the spatial domain $[0, 2] \times [0, 1]$. 
\begin{equation}\label{eqn:Double_Gyre} 
\begin{aligned}
    \psi(x, y, t) &= A sin(\pi f(x, t)) sin(\pi y) \\
    f(x, t) &= a(t)x^2 + b(t)x \\
    a(t) &= \varepsilon sin(\omega t) \\
    b(t) &= 1 - 2\varepsilon sin(\omega t) \\
    \text{where}~A &= 0.1,~\omega = pi/5~\text{and}~\varepsilon = 0.25 \\
\end{aligned}
\end{equation}
\begin{figure}[]
    \centering
    \subfigure[Glyph-based visualization of the velocity field at time 0.]
    {\label{fig:data_0}\includegraphics[width=0.48\textwidth]{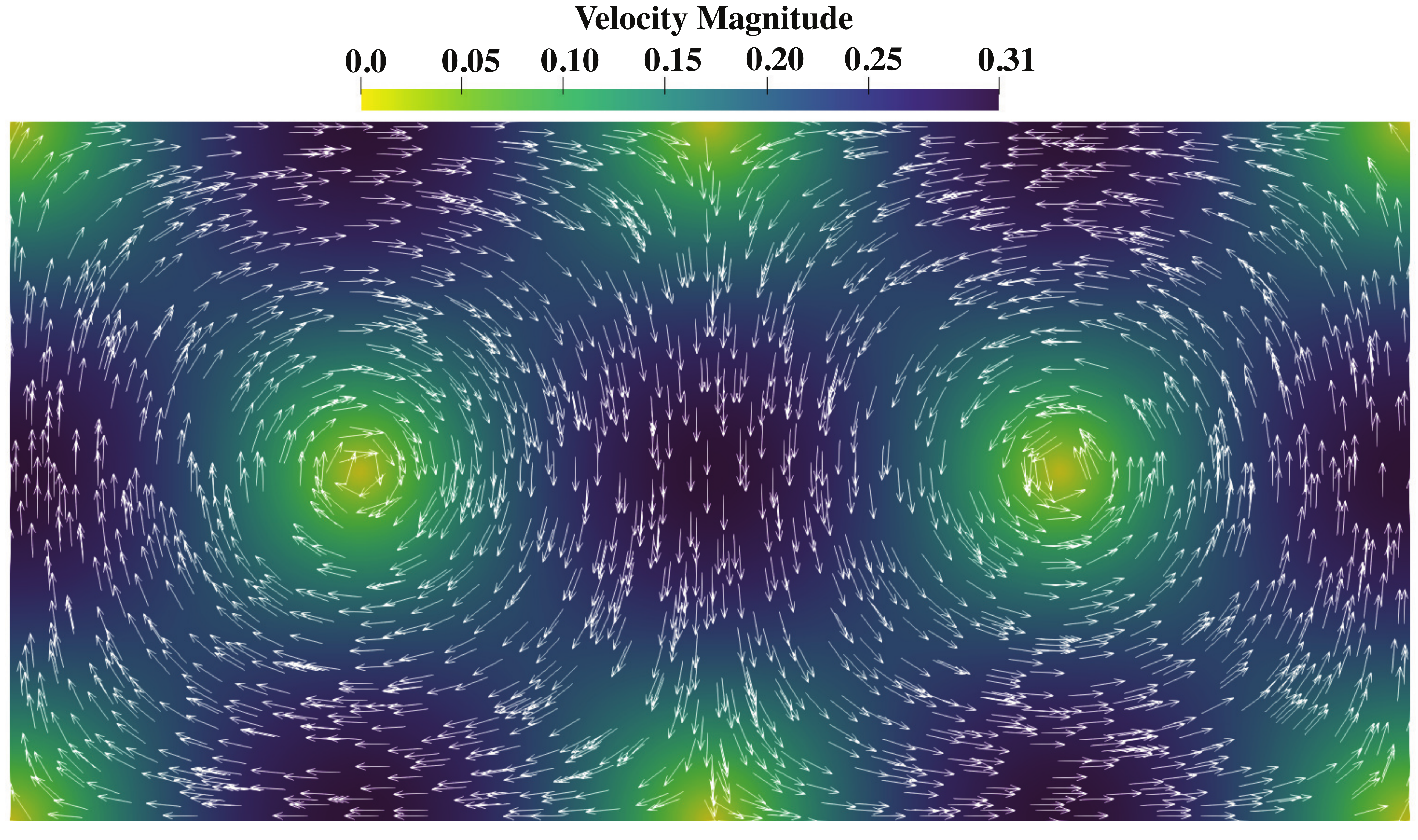}}
    \quad
    \subfigure[Forward FTLE scalar field computed over 1,000 cycles.]
    {\label{fig:data_1}\includegraphics[width=0.48\textwidth]{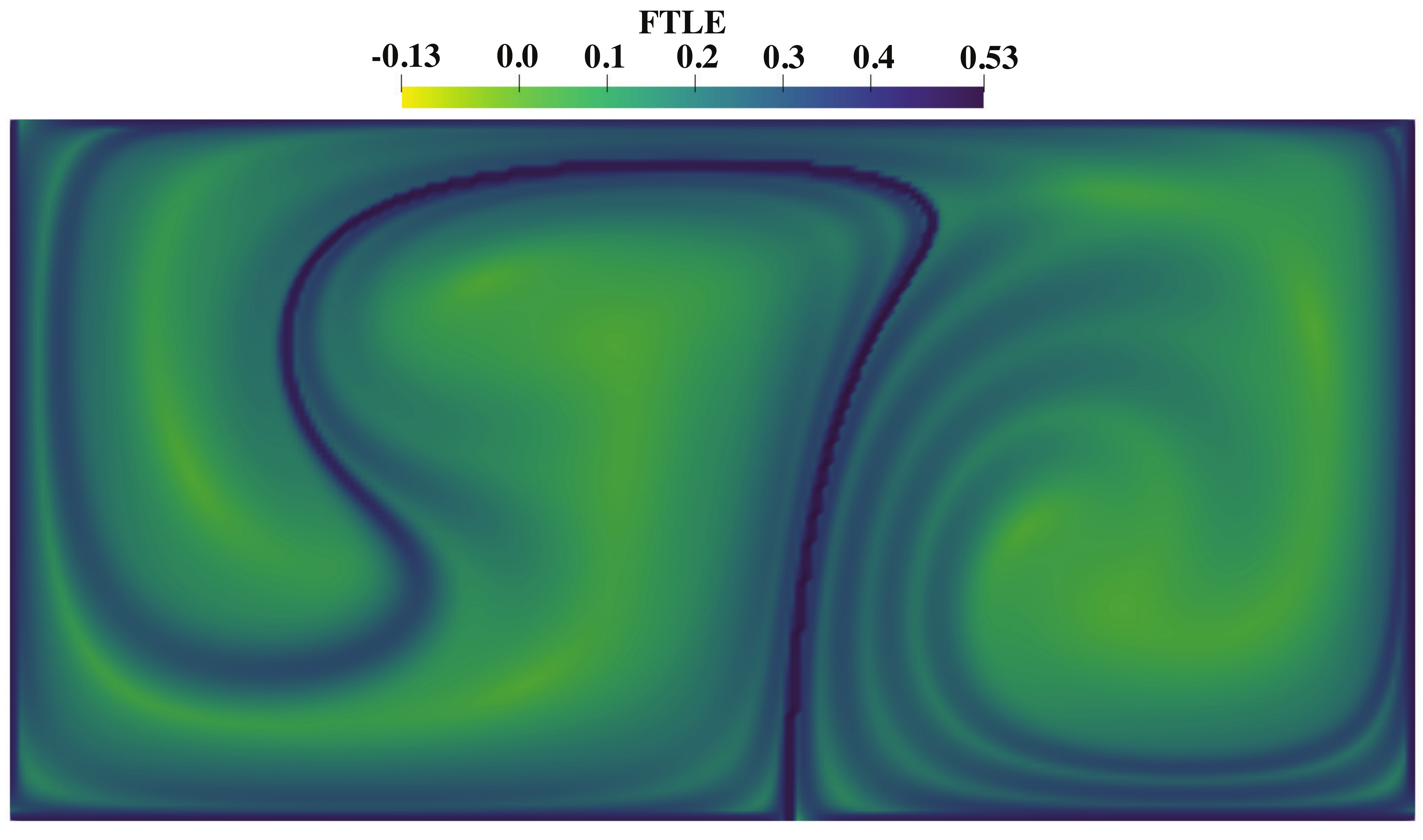}}
    \caption{\label{fig:data_description}Visualizations of the Double Gyre data set showing the two counter-rotating gyres (Figure \ref{fig:data_0}) and the Lagrangian coherent structures as approximated by the ridge of the finite-time Lyapunov exponent (FTLE) scalar field (Figure \ref{fig:data_1}).}
\end{figure}

Our training data generation process used the analytical solution~(Equation~\ref{eqn:Double_Gyre}) for particle advection during Lagrangian flow map computation. 
We show the velocity field at time 0 (Figure \ref{fig:data_0}) and the FTLE (Figure \ref{fig:data_1}) of the Double Gyre data set.
The ridges of the FTLE scalar field are used to approximate Lagrangian Coherent Structures in the flow.
We extended the 2D Double Gyre data sets to 3D by adding the same z-axis to every seed. 
%Prior studies have demonstrated the use of sparse Lagrangian flow maps to accurately reconstruct flow fields, extract features, and offer significant data reduction.
%
%For example, a training data set containing 5,000 seeds with a file cycle interval of 30 over Y cycles of a three X-dimensional vector field required only 4.1MB when stored in the NPY file format. 
%
The size of training data sets increases linearly with a larger number of seeds and shorter intervals. 
In our experiments, the minimum and maximum sizes of the reduced Lagrangian representation training data were $2.6$MB and $24.2$MB, respectively. 
We did not observe significant improvements of accuracy using more training data for this data set.
We generated all the training data sets using a desktop equipped with an Intel(R) Xeon(R) W-3275M CPU ($56$ cores; $256$GB memory) and one NVIDIA Titan RTX GPU. 
We computed the particle trajectories of the Lagrangian flow maps in parallel using the TBB library (\cite{Advanced_HPC_Threading}).

%In addition, we train a model using a real-world Jet simulation~\mengjiao{cite} for the post hoc reconstruction (Section \ref{sec:reconstruction}). 
%
%This Jet data set contains 730 cycles in total, with each cycle consisting of a uniform grid with dimension $[128 \times 256 \times 128]$. 
%
%The particle advection process is done by VTK-m~\cite{moreland2016vtk} with the 4th order Runge-Kutta method (RK4) when generating the training data sets. 

\subsection{Evaluation of Seeding Strategy and Hyperparameters Settings}
\label{sec:hyper-parameters}
Our model was implemented using the Pytorch library (\cite{NEURIPS2019_9015}) and trained on dual RTX 3090s GPUs.
We considered two methods of extracting training data sets~(Section \ref{sec:training_data}): $Lagrangian_{long}$ and $Lagrangian_{short}$.
We studied the impact of seeding strategy as well as the learning rate and batch size for each flow map extraction approach.
%
%In addition to varying generation processes, the inference process when using the model trained by data from these two approaches also varies.
%
%When using $Lagrangian_{long}$, interpolations are performed by always considering the new seed start location at $t_0=0$.
%
%The end location inferred by the model results from the provided start location and the queried file cycle.
%
%In contrast, when using $Lagrangian_{short}$, new particle trajectories are ``stitched'' together by advancing the new seed across intervals.
%
%Here, the inference is performed by considering the location of the seed particle at the previous file cycle and the target file cycle.
%
%Since every inference except the first uses previously inferred results, errors might propagate along new trajectories when using $Lagrangian_{short}$~\cite{hummel2016error, sane2019interpolation}.
%
%We refer to \textit{Global Error} as updating every inference using previous inferred results and \textit{Local Error} 
%as the absolute error introduced by the model in following results.   

\subsubsection{Seeding Strategy}
\label{sec:seeding_strategy}

\begin{figure}[]
    \centering
    \subfigure[$Lagrangian_{long}$ tests.]
    {\label{fig:seeding-long}\includegraphics[width=\textwidth]{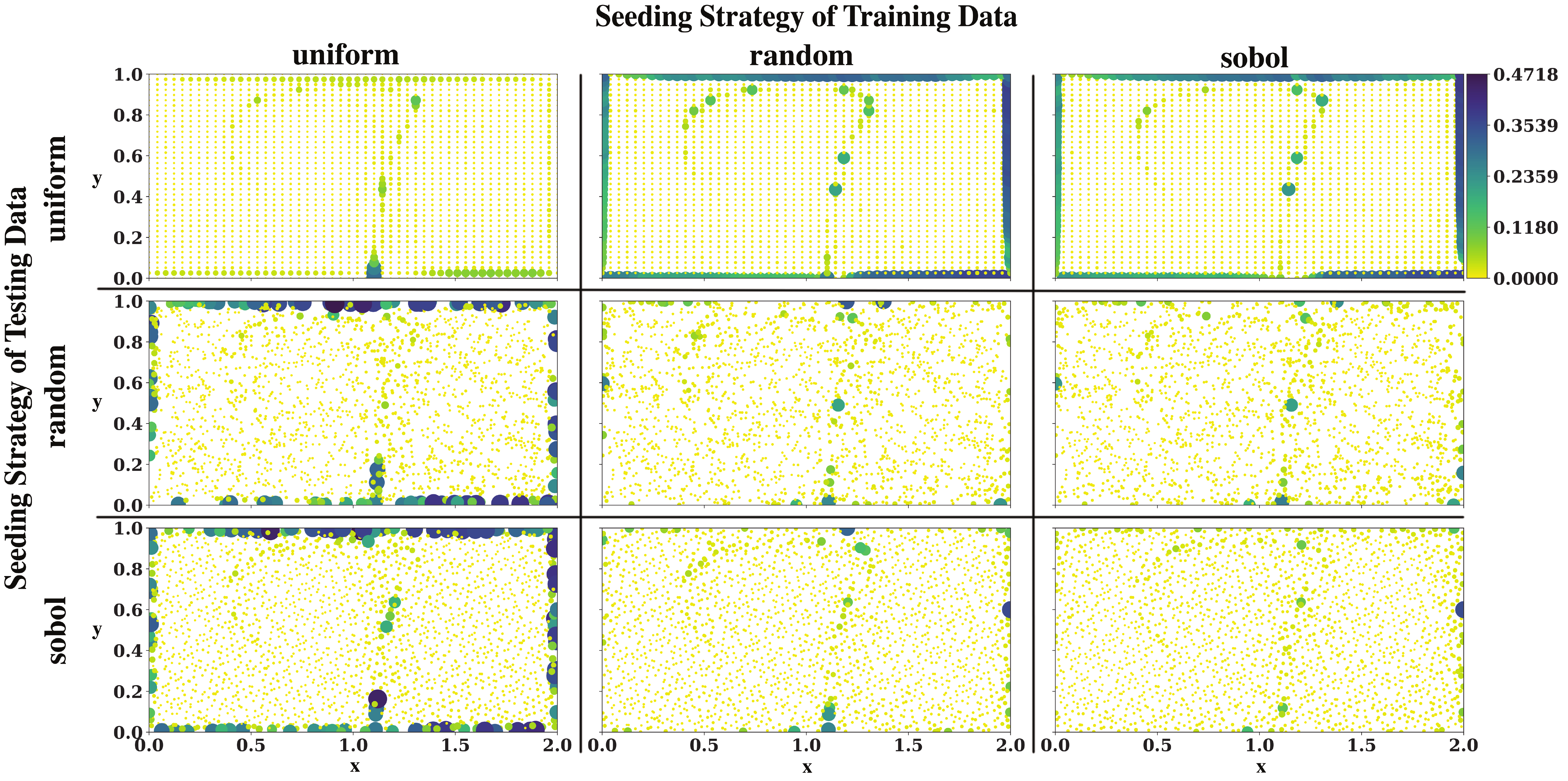}}
    \subfigure[$Lagrangian_{short}$ tests.]
    {\label{fig:seeding-short}\includegraphics[width=\textwidth]{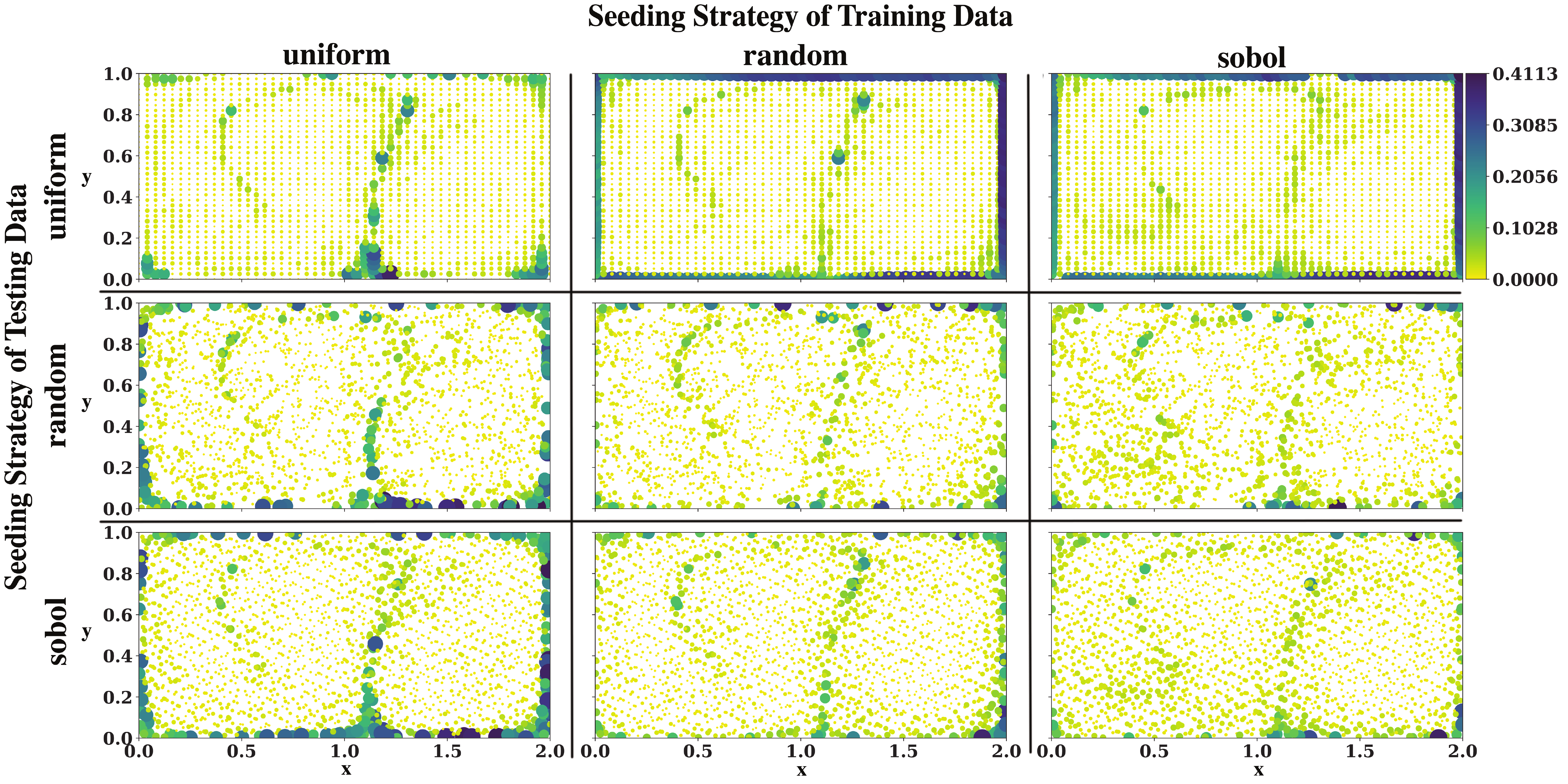}}
    \caption{\label{fig:seeding_strategy}Visualization of the errors mapped to the particle trajectory start location for three sampling strategies applied to generate both training and testing data sets.
Figures~\ref{fig:seeding-long} and \ref{fig:seeding-short}, show results for the $Lagrangian_{long}$ and the $Lagrangian_{short}$ flow map extraction strategies, respectively.
The columns (left to right) represent $uniform$, $random$, and $sobol$ sampling for training seeds. 
The rows (top to bottom) represent $uniform$, $random$, and $sobol$ sampling for testing seeds.
%
%Error maps are evaluated by applying three seeding strategies to training data sets and testing data sets.  
%    In Figure \ref{fig:seeding-long} to \ref{fig:seeding-short}, training seeds are placed using pseudorandom number sequence ($random$), 
%    Sobol quasirandom sequence ($sobol$) and uniformed grid ($uniform$) from left to right. 
For example, in Figure \ref{fig:seeding-long}, column 1 row 3 shows the result of using uniform seeding for training sample generation 
and sobol seeding for testing reconstruction when using the $Lagrangian_{long}$ strategy.
Each figure shows the spatial domain $[0, 2] \times [0, 1]$.
The testing data contains 2,000 seeds for $random$ and $sobol$, and uses a $[50 \times 40]$ grid for $uniform$.
The error is measured by aggregated along the trajectories (Equation \ref{eq:measure}) and is encoded in the visualization using the color and area of each circle mark.
%
%The seeds whose error is in the top 1\% are as outliers and removed for analysis.
Overall, we find the $sobol$ or the Sobol quasirandom sequence strategy performs the best as a training and testing data sampling strategy across both flow map extraction approaches. 
However, we find the studied strategies can result in poor extrapolation for particles placed on the boundary.}
\end{figure}

To generate training data, we evaluated three seed placement strategies: (1) seeding along a uniform grid ($uniform$), (2) seeding using a pseudorandom number 
sequence ($random$), and (3) seeding using a Sobol quasirandom sequence ($sobol$). 
For this experiment, we sampled the time-varying Double Gyre vector field domain using 2,000 seeds and a fixed file cycle interval of 30.
All models were trained with a batch size of 200 and a learning rate of $0.001$.
%
%For each of the three seed placement strategies, we placed 2,000 seeds in the domain to generate training data. 
%
For the uniform sampling experiment, we used a $[50 \times 40]$ grid.
Further, besides applying these three seed placement strategies to generate training data sets, we also considered the strategies for testing seeds.
Figure~\ref{fig:seeding_strategy} presents error maps produced by various combinations of seed placement strategies for training and testing data, as well as outcomes considering two flow map extractions strategies.
Comparing error maps evaluated by using $Lagrangian_{long}$ for sampling time-varying vector field (Figure \ref{fig:seeding-long}), we found that the Sobol quasirandom sequence ($sobol$) was slightly better than the pseudorandom number sequence ($random$). 
They both produced more accurate results for the testing seeds that were not on the boundary. The uniform seeding ($uniform$) was more accurate only when the testing seeds were also uniform.
Moreover, the Sobol quasirandom sequence ($sobol$) performed better than the pseudorandom number sequence ($random$) 
when sampling the time-varying vector field using $Lagrangian_{short}$, and they were both better than the uniform 
seeding ($uniform$) (Figure \ref{fig:seeding-short}) except for seeds on the boundary.
We chose the Sobol quasirandom sequence ($sobol$) as the seeding strategy in all our following experiments. 
%and set a boundary offset of $0.05$ to for a fair comparison. 
%
%Further work is required to improve this seed strategy and reconstruct new trajectories seeded on the boundary. 
Further work is required to identify sampling strategies that optimize the quality of the training data.

\subsubsection{Learning Rate and Batch Size}

\begin{figure}[]
    \centering
    {\includegraphics[width=\textwidth]{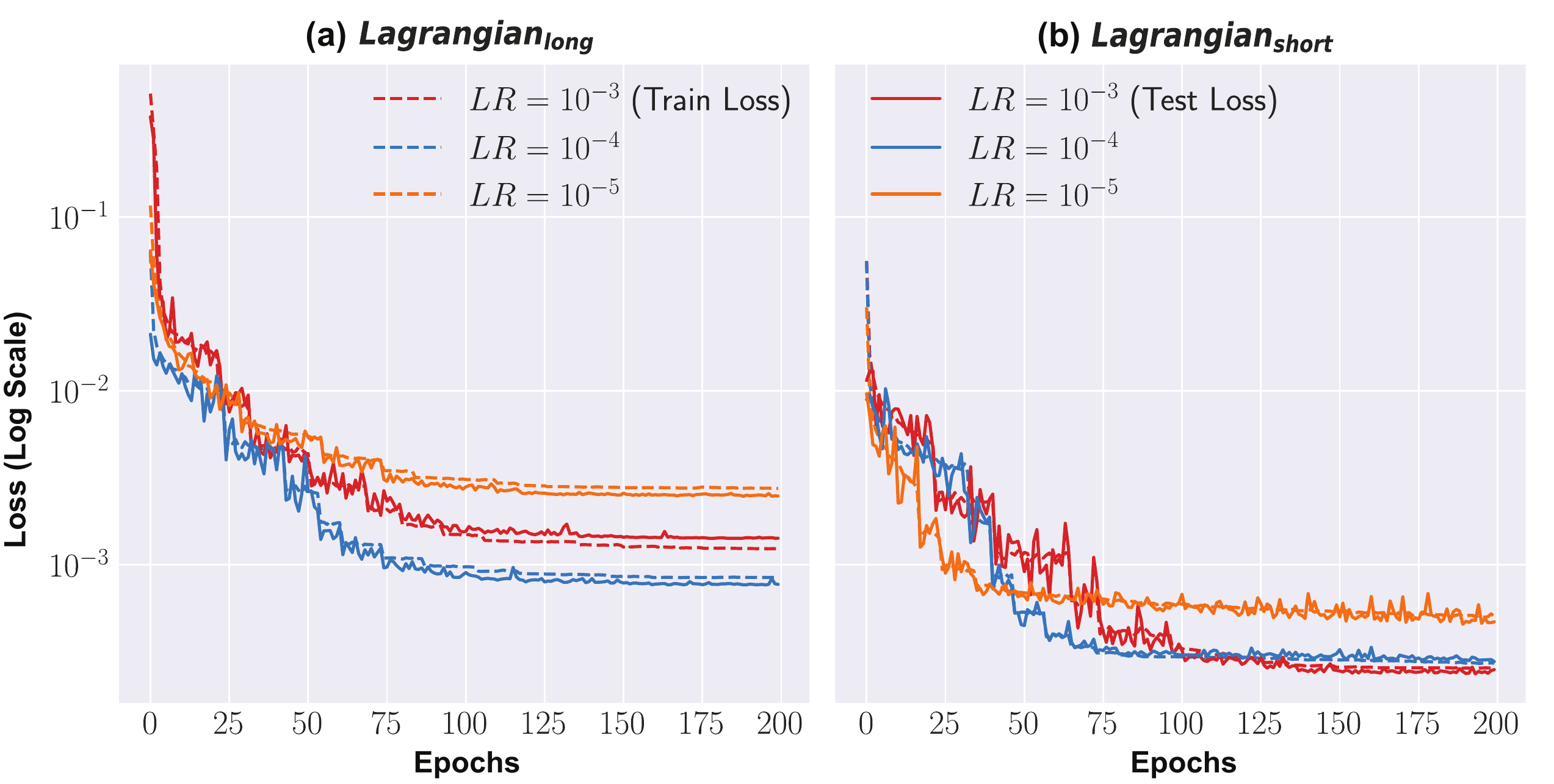}}
    \caption{\label{fig:loss_epoch}Loss versus epoch plots considering multiple learning rates for the two flow map extraction strategies. We use the learning rates $10^{-3}$, $10^{-4}$, and $10^{-5}$. 
%for $Largrangian_{long}$ and $Lagrangian_{short}$ approaches. 
The training data set is generated by placing 5,000 seeds using the $sobol$ method and file cycle interval is set to 30. 
%While the model trained with $Lagrangian_{long}$ training data shows a decrease in loss using larger learning rates~($10^{-3}$, $10^{-4}$), the model trains better using a smaller learning rate~($10^{-4}$, $10^{-5}$) when $Lagrangian_{short}$ training data is used.
}
\end{figure}
The learning rate is a critical hyperparameter for a deep neural network. 
We examined four settings of the learning rate: $10^{-2}$, $10^{-3}$, $10^{-4}$, and $10^{-5}$ for $Lagrangian_{long}$ and $Lagrangian_{short}$. 
For all experiments, the training data sets were generated with 5,000 seeds and a file cycle interval of 30 using the $sobol$ seed placement method with the Double Gyre data set.
The batch size was set to 200.  
The learning rate of $10^{-2}$ resulted in the model failing to converge; therefore, we did not use it for comparison. 
We found the learning rates of $10^{-3}$ and $10^{-4}$ were better for our model when the training data sets were generated using the $Lagrangian_{long}$ flow map extraction strategy~(Figure~\ref{fig:loss_epoch}(a)).
The learning rates of $10^{-3}$, $10^{-4}$, and $10^{-5}$ resulted in a similar loss when the model was trained using data sets generated using the $Lagrangian_{short}$ approach~(Figure \ref{fig:loss_epoch}(b)).

\begin{figure}[]

    \centering
    \subfigure[5,000 seeds.]
    {\label{fig:lr-a}\includegraphics[width=\textwidth]{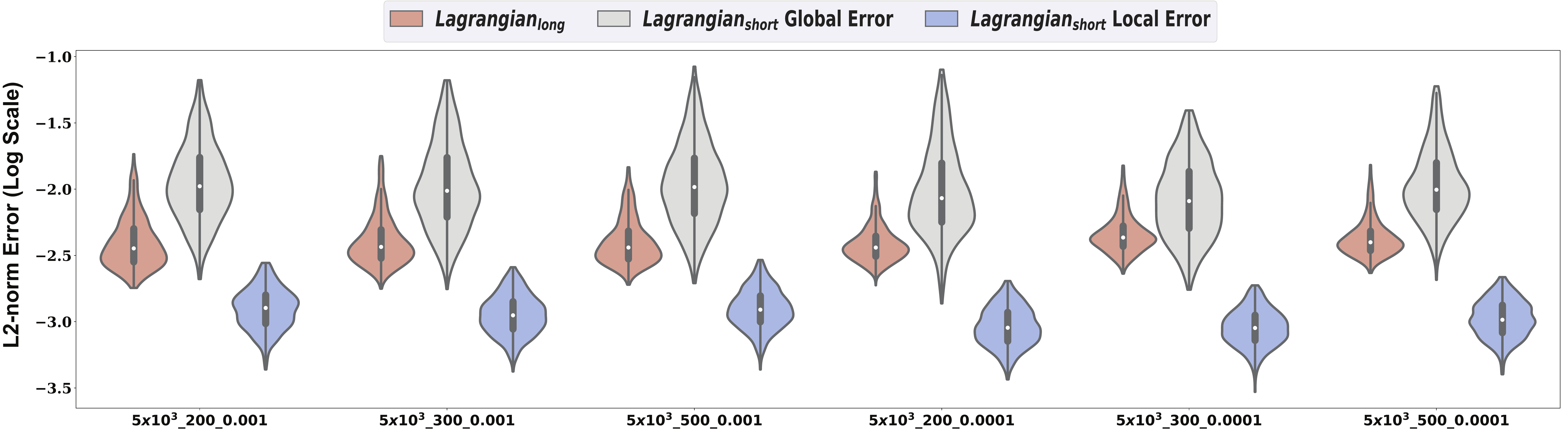}}
    \subfigure[10,000 seeds.]
    {\label{fig:lr-b}\includegraphics[width=\textwidth]{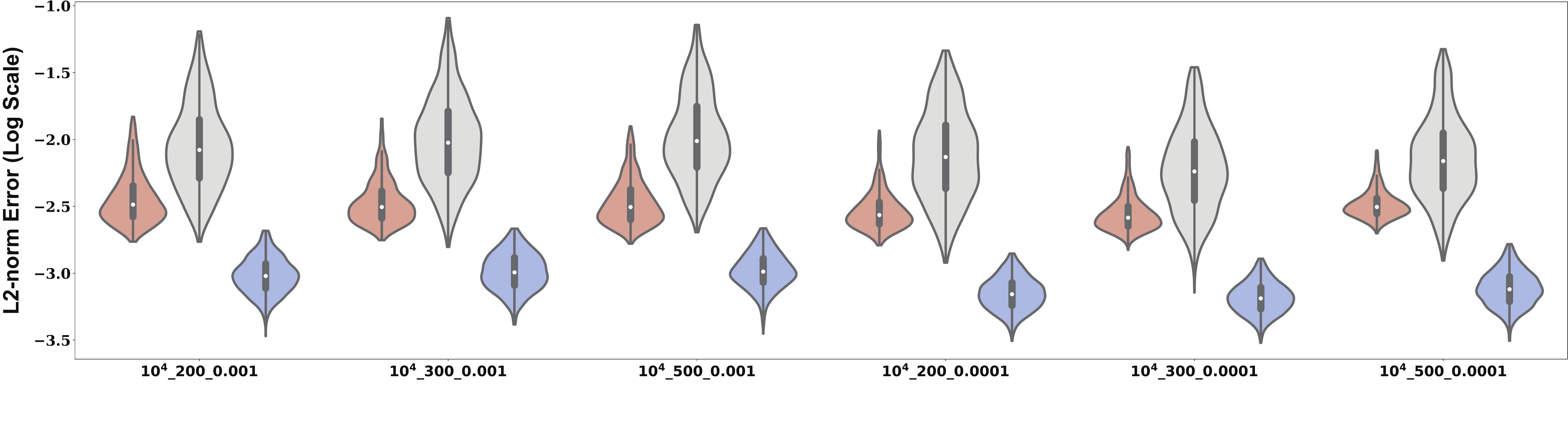}}
    \caption{\label{fig:lr}The error plot evaluated various combinations of the learning rate and the batch size for $Lagrangian_{long}$ and $Lagrangian_{short}$ approaches. 
    The errors are evaluated over 2,000 seeds and aggregated along the trajectories ((Equation \ref{eq:measure})). 
    The labels on the y-axis use $10^X$ format to show the error. 
    We use the format $N\_B\_LR$ to label each set of tests, where $N$ is the number of seeds, $B$ is the batch size, and $LR$ is the learning rate.
    The top 1\% of errors in each experiment are treated as outliers and have been removed for analysis.
    A batch size of 200 with the learning rates of $10^{-3}$ and $10^{-4}$ are optimal for training data sets with 5,000 seeds and 10,000 seeds, respectively, using the $Lagrangian_{long}$ approach. 
    A batch size of 300 with the learning rate $10^{-4}$ is optimal for the $Lagrangian_{short}$ approach.} 
\vspace{-1em}
\end{figure}

To identify the optimal combination of batch size with the learning rates of $10^{-3}$ and $10^{-4}$, we conducted a set of experiments.
Our experiments considered three options for batch size, two options for total number of training samples, and both flow map extraction strategies ($Lagrangian_{long}$ and $Lagrangian_{short}$).
Figure~\ref{fig:lr}) presents violin plots of the error for reconstructed trajectories.
Although we found the choice of learning rate and flow map extraction strategy could significantly impact accuracy, varying the batch size did not result in a significant change of accuracy for a fixed learning rate and flow extraction strategy.
%
% \sudhanshu{Mengjiao check if the previous statement is true.}
%We use the optimal hyperparameter settings in our following experiments.  

% In Figure \ref{fig:lr-a}, L01 - L03: $N=0.165M$, $lr = 0.001$ and $b = 200, 300 and 500$; 
%     L04 - L06: $N=0.165M$, $lr = 0.0001$ and $b = 200, 300 and 500$; L07 - L09: $N=0.33M$, $lr = 0.001$ and $b = 200, 300 and 500$; L10 - L12: $N=0.33M$, $lr = 0.0001$ and $b = 200, 300 and 500$.
% We refer to the number of training samples as $N$, the learning rate as $lr$, and the batch size as $b$. 
\subsection{Network Training and Inference}
\label{sec:network_training}
%
%
% We applied the Adam optimizer with an initial learning rate of $10^{-5}$ and employed a learning rate scheduler to adjust the rate based on the validation loss.
% %
% Readers can refer to Section \ref{sec:training_process} for more details. 
%
Table~\ref{tab:performance} reports time spent training the model, memory consumption for saving the trained model, and the inference time to generate new trajectories with the trained model. 
%
% Note the number of training samples (\#Samples$_{long}$ and \#Samples$_{short}$) is not equal to the product of the number of seeds and file cycles for the ABC data set.
%
%Differs from the Double Gyre, where particle trajectories may be outside the domain after some cycles for ABC data sets.
%
%These trajectories are useless for presenting the flow behaviors that we do not add into the training samples.
%
%To account for particles exiting the domain as long particle trajectories are computed, the number of initial seeds used for the ABC data set for $Lagrangian_{long}$ is larger than the number of seeds used for $Lagrangian_{short}$.
% As a consequency of particles exiting the domain as long particle trajectories are computed, the number of samples for $Lagrangian_{long}$ is smaller than the number of samples used for $Lagrangian_{short}$.
%
%Tracing long trajectories results in more chances to be out of the domain and reduces more training samples. 
%
As expected, the training time increased linearly with the number of training samples for both approaches. 
%
% Further, the ABC data set requires longer for training than the Double Gyre because of the increased complexity of learning an additional spatial dimension of the flow field.
%increased complexity of the flow field. 
%
The storage cost for saving the trained model, irrespective of the data set or number of training samples, was fixed. 
Based on the network's parameters, the trained models required the same memory size of 10.5 MB.
%
% We expect the model can be trained using data from more complex, turbulent, and 3D flow fields.
%
% However, verification as well as understanding impacts flow field complexity on network training and performance requires a future in-depth investigation. 
However, verification as well as understanding the impacts of complex, turbulent, and 3D flow fields on network training and performance requires future in-depth investigation. 
That said, considering the network's parameters are independent of the complexity of the flow field, we expect our method to scale and be used to reduce the memory footprint of large-scale high-resolution Lagrangian representations of time-varying vector fields.
%
%The trained models use the same space size (10.5 MB), which indicates that the network's parameters are independent of the complexity of flow behavior. 
%
%Thus, we expect our method to scale can be used to reduce the memory footprint of high-resolution vector fields.
%
%For example, a 3D vector field defined on a uniform $256^{3}$ mesh over 1000 cycles using floating-point precision, requires over 200 GBs in an Eulerian representation (full spatial and temporal resolution stored) and approximately 8 GBs in a Lagrangian representation (full spatial resolution and an interval of 25 cycles).
%
%, e.g., a 3D ABC flow map with resolution $256^3$ needs 13.69 GB with file cycle interval 30 using the NPY file format. 
%
An important consequence of a small memory footprint is the reduced cost of two seconds to load the entire model, thus alleviating the system from expensive I/O for loading data during exploratory visualization. 
Further, our results show parallel inference of 2,000 trajectory with 20 locations interpolated to approximate each curve costs 0.38s using the same machine as for generating training data sets. 

\begin{table}[]
    \caption{\label{tab:performance}Network training and computational performance results.
We present the number of seeds (\#Seeds), file cycle interval (Interval), number of training samples (\#Samples), the training time (Train), trained model storage space (Model), and the inference performance (Inference) details of our experiments.
The training time is measured for 100 epochs and increases linearly with the number of training samples. 
Importantly, our method costs 10.5MB memory for storing the trained model regardless of the number of training samples, potentially significantly reducing the storage space for large-scale time-varying vector fields.
The inference time for 2,000 new particle trajectories interpolated across 1,000 cycles is presented.
The interpolation of each location along a particle trajectory advances the particle by the length of the file cycle interval.}
\begin{adjustbox}{width=0.9\textwidth, center}
    % \centering
    \begin{tabular}{cccccc}
    \hline
    \textbf{\#Seeds} & \textbf{Interval} & \textbf{\#Samples (M)} & \textbf{Train (hrs)} & \textbf{Inference (s)} & \textbf{Model (MB)} \\ \hline
    5,000            & 30                & 1.65                   & 0.44                 & 0.54                   & 10.5                \\
    10,000           & 30                & 3.30                   & 0.86                 & 0.54                   & 10.5                \\
    10,000           & 50                & 2.00                   & 0.55                 & 0.38                   & 10.5                \\ \hline
    \end{tabular}
\end{adjustbox}
\end{table}

\subsection{Impact of Number of Seeds}
\label{sec:num_seeds}
We evaluated the impact of the number of seeds on the performance of our model qualitatively and quantitatively.
We used a fixed file cycle interval of 30 for all training data discussed in this section.
We created training data sets with four options for number of seeds, 5,000, 10,000, 15,000, and 20,000, for the $Lagrangian_{long}$ 
and $Lagrangian_{short}$ approaches.
To evaluate the accuracy of the reconstruction, 2,000 random particles were seeded in the domain. 
To avoid extrapolation errors due to our use of the $sobol$ seeding strategy for training data generation~(Section~\ref{sec:training_data}), 
we used a boundary offset of $0.05$ to prevent test seeds from being placed exactly on the boundary.

\begin{figure}[]
    \centering
    \subfigure[Particle trajectory reconstruction error mapped to particle start location when varying the number of seeds used to generate training data.]
    {\label{fig:1a}\includegraphics[width=\textwidth]{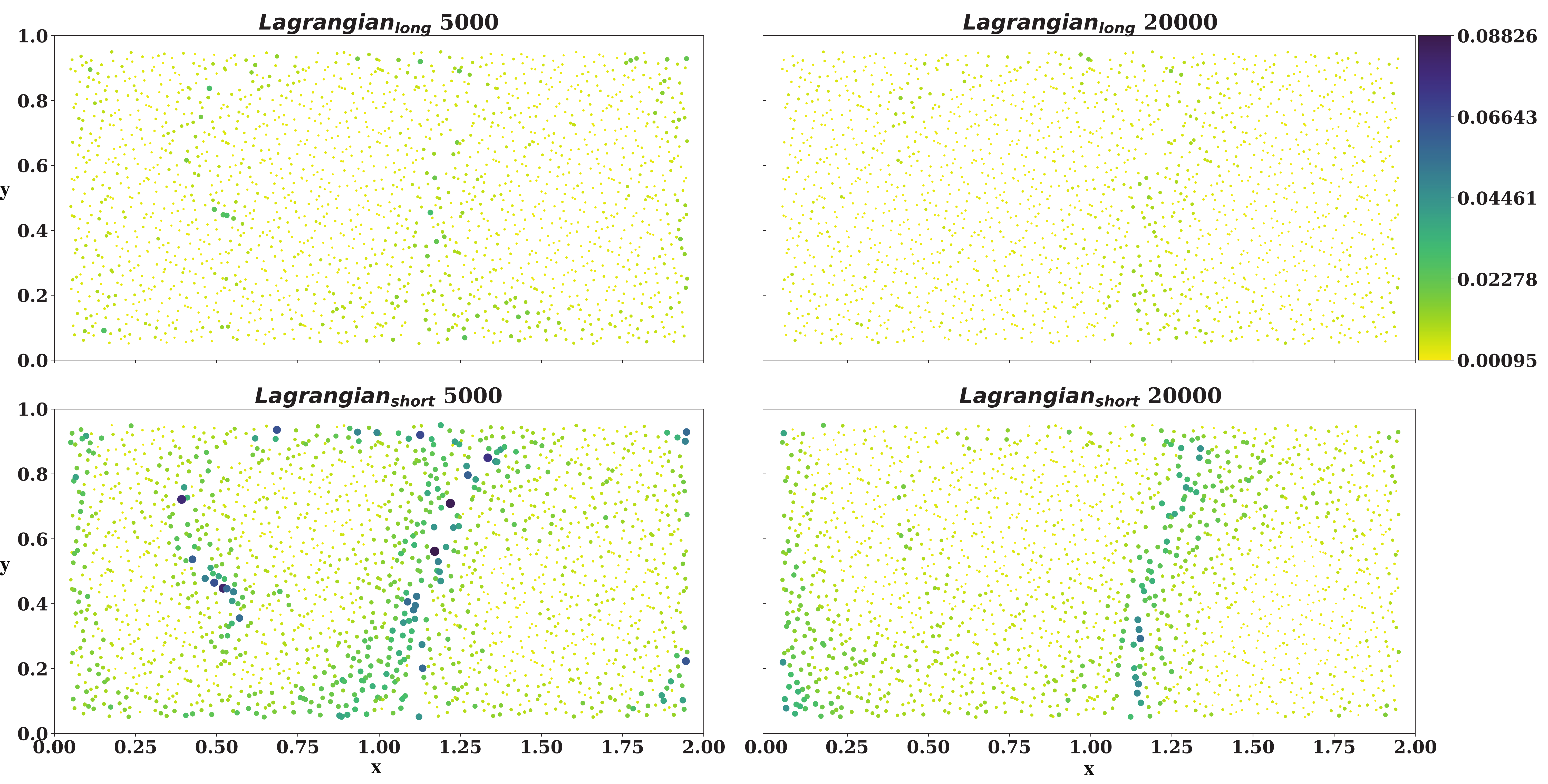}}
    % \quad
    % \subfigure[Double Gyre - Lagrangian\_short - File cycle interval 30]
    % {\label{fig:1b}\includegraphics[width=\textwidth]{figures/error_maps/short_num_seeds_error_map.pdf}}
    \subfigure[FTLE scalar field derived using trajectories inferred from the model.]
    {\label{fig:1b}\includegraphics[width=\textwidth]{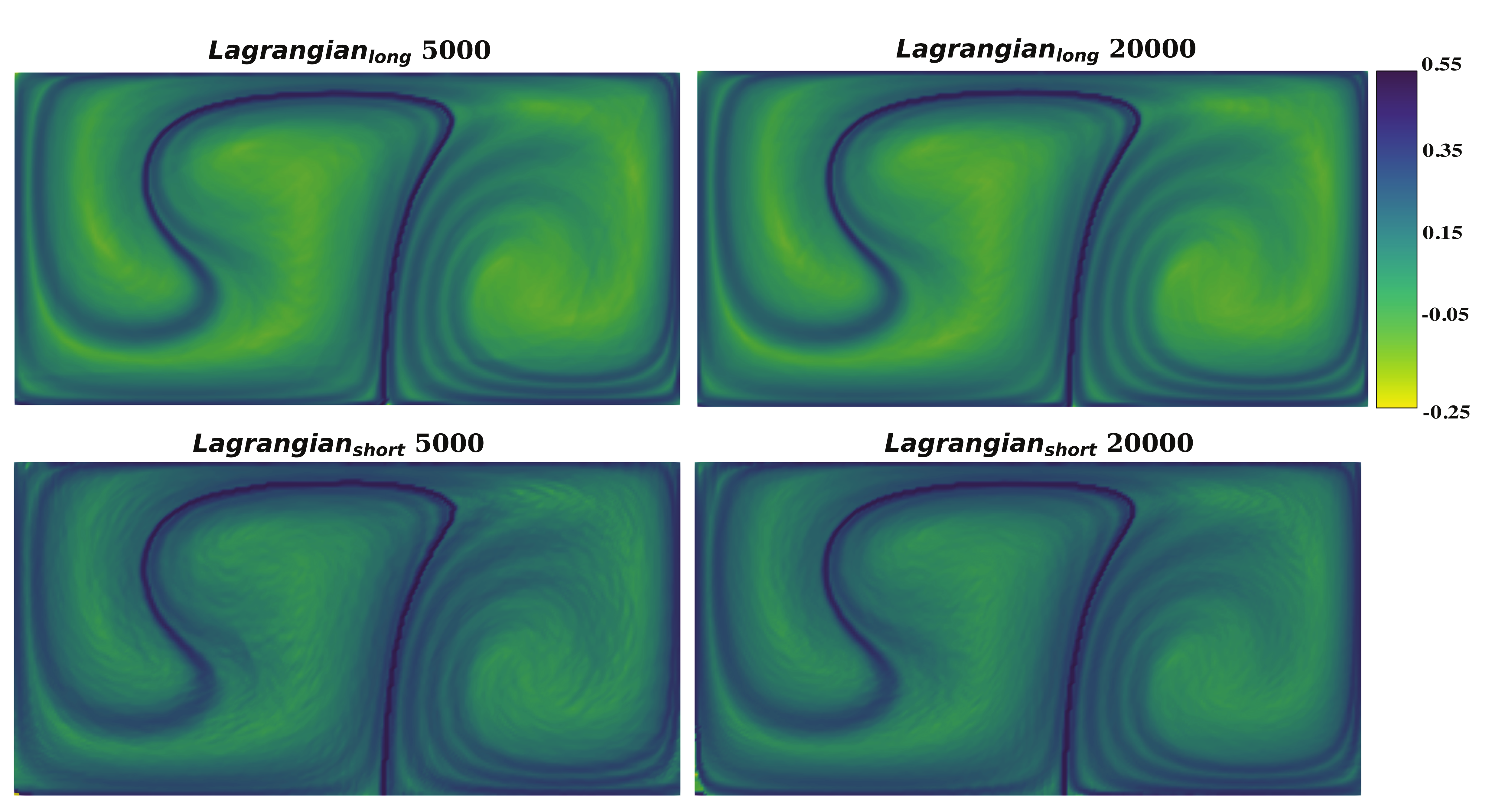}}
    \caption{\label{fig:num_seeds_error_map} Visualization of particle trajectory reconstruction error mapped to particle start locations~(\ref{fig:1a}) 
    and the corresponding FTLE scalar fields derived from trajectories inferred by the model~(\ref{fig:1b}), when varying the number of seeds used to generate training data. 
%
%
%The model's performance with varying numbers of seeds is shown as error maps analyzing the model's performance according to the seeds' location in the domain. 
%
Each figure shows the spatial domain $[0, 2] \times [0, 1]$.
The models are trained with a file cycle interval of 30 and the best combination of hyperparameter settings identified in Section~\ref{sec:hyper-parameters}.
We evaluate reconstruction error using 2,000 seeds visualized as circle marks in \ref{fig:1a}. 
The color and radius of the circles encode the error aggregated along the trajectories (Equation \ref{eq:measure}). 
The top 1\% of errors are treated as outliers and have been removed for analysis from each experiment.
The FTLE is calculated by placing a uniform grid with size $[256 \times 128]$. 
The model's performance is related to the flow behavior in the domain, and reconstruction errors are higher in regions with greater separation, notably for the $Lagrangian_{short}$, which suffers from error propagation.}
\end{figure}

\begin{figure}[]
    \label{fig:2b}
    {\includegraphics[width=\textwidth]{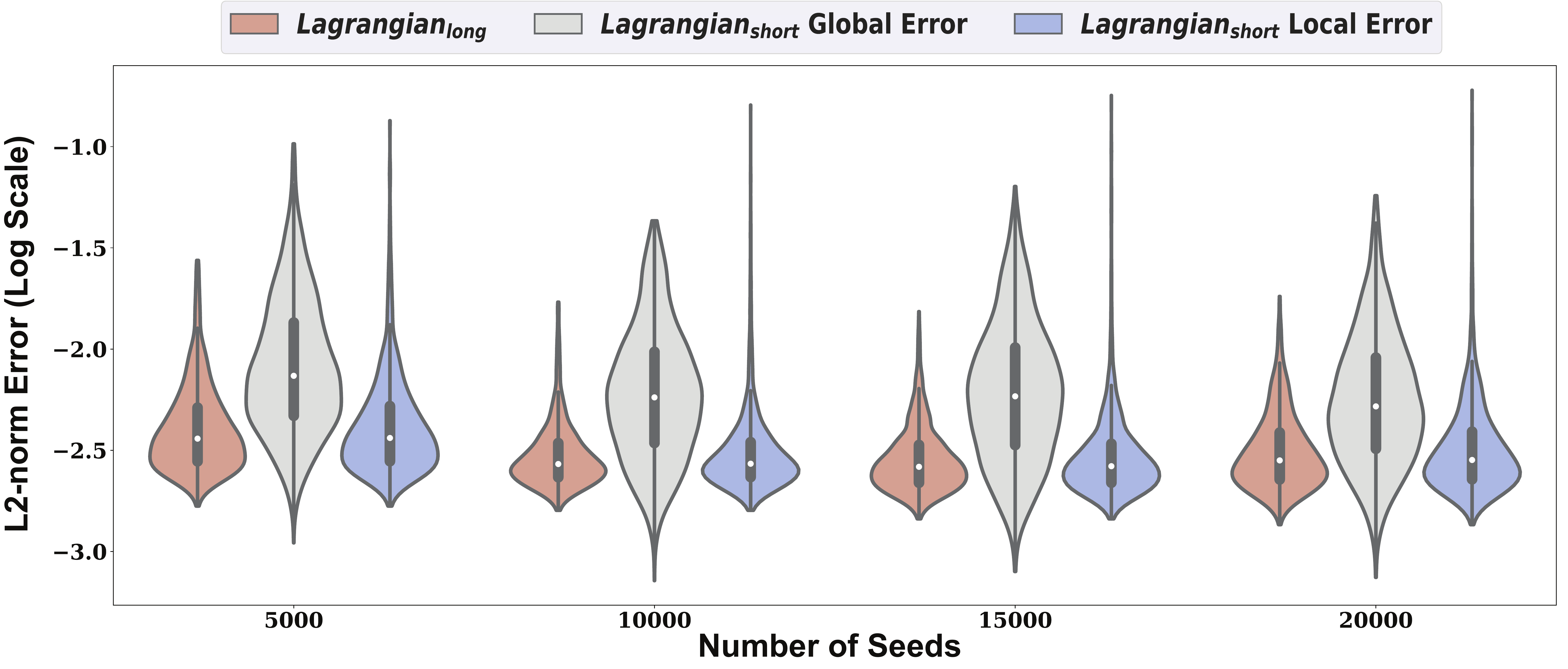}}
    \caption{\label{fig:num_seeds_violin}
Violin plots of inference error evaluated for models trained using data generated with varying the number of seeds.
    The errors are calculated along the trajectories using Equation \ref{eq:measure}. The labels on the y-axis use $10^X$ format to show the error. The error is shown as a distribution using violin plots with the minimum, maximum, and median errors. The evaluation is performed using 2,000 random test seeds.
    The top 1\% of errors are treated as outliers and have been removed for analysis from each experiment.
Our results indicate the inference accuracy can improve from increasing the number of seeds used to train the model.}
\end{figure}

% [FTLE computed using inference results of two approaches and the ground truth]
\begin{figure}[]
    \includegraphics[width=\textwidth]{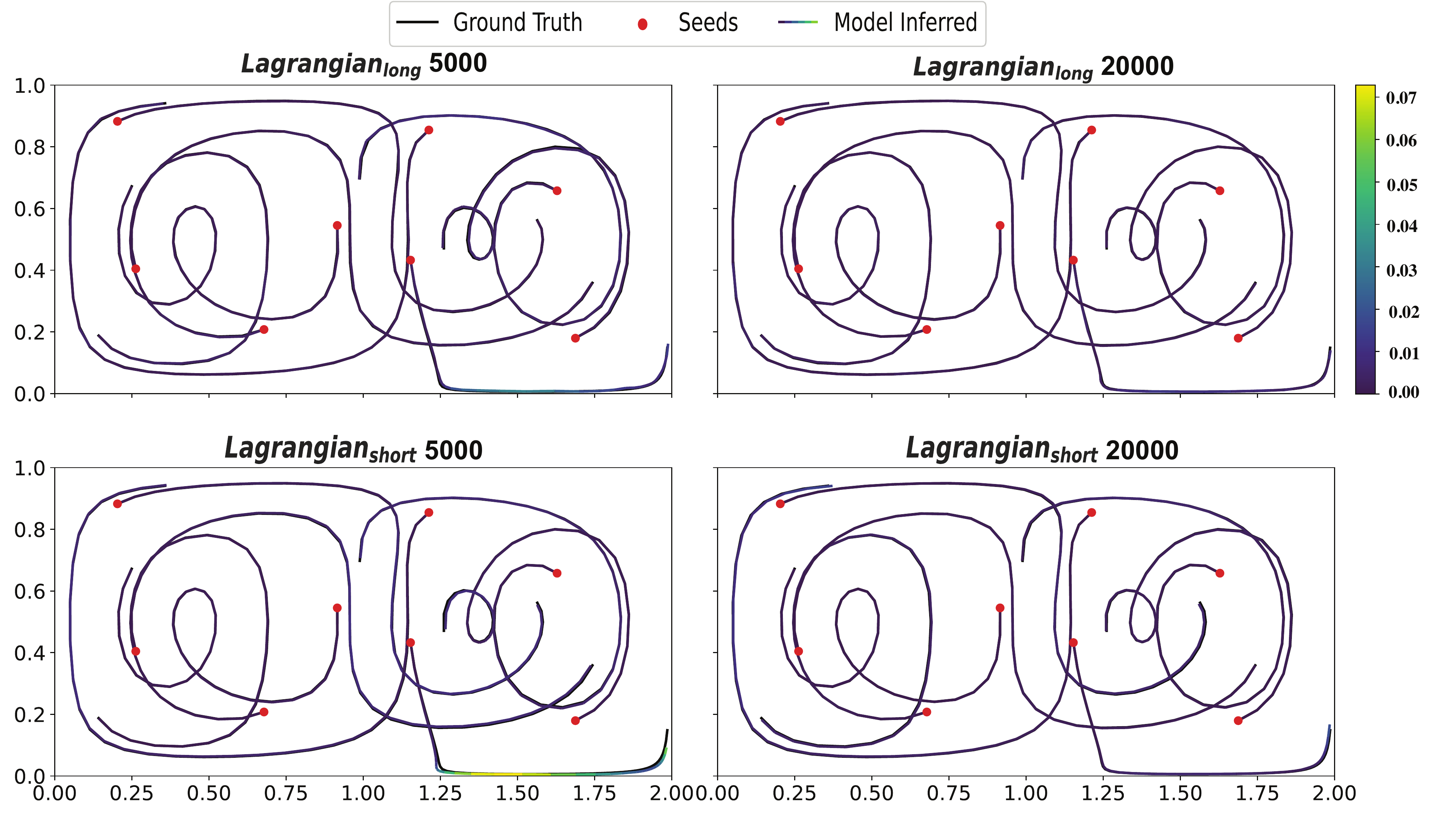}
    
    % \subfigure[Double Gyre - Lagrangian\_short - File cycle interval 30\label{fig:3b}]
    % {\includegraphics[width=\textwidth]{figures/seed_rake/num_seeds_short_seed_rake.pdf}}
    \caption{\label{fig:num_seeds_trajectories}Visualization of inferred trajectories and the ground truth for the Double Gyre with different numbers of seeds used to train the model.
    The seeds were randomly placed using Sobol seeding strategy.  
    The colors of model inferred trajectories indicate the distance between the model inferred end location and the ground truth. 
    In nearly all cases, our trained models can reconstruct trajectories almost visually identical to the ground truth.
}
\end{figure}

In Figure~\ref{fig:num_seeds_error_map}, we report the error map as well as the FTLE derived from using various configurations for training data generation.
%
% The ridges of the FTLE scalar field are used to approximate Lagrangian Coherent Structures in the flow.
%
The result highlighted the relation of the trained model's performance and flow features in the domain. 
The error for each trajectory was measured using Equation \ref{eq:measure}.
% AEDR metric proposed by \cite{ren2020uncertainty}. 
%
We observed reconstruction errors were higher in regions with greater separation in the flow field, i.e., regions with higher FTLE values.
Moreover, for both $Lagrangian_{long}$ and $Lagrangian_{short}$, the error maps confirmed that increasing the number of seeds could increase the inference accuracy.
%
% In addition, we measure the individual distance between the model-generated results to the ground truth~(Figure \ref{fig:num_seeds_violin}).
In addition, we visualized the distribution of errors for the model-generated results in comparison to the ground truth~(Figure \ref{fig:num_seeds_violin}).
We observed a decreasing median error as the number of seeds used to sample the domain increased. 
%There is a clear trend of decreasing mean and median error by increasing the number of seeds in the domain.
%
However, the reduction in error was less after 10,000 seeds. 
Further, the models trained with $Lagrangian_{short}$ data sets showed greater global error due to local error propagation during reconstruction of new trajectories. 
In the derived FTLE fields in Figure~\ref{fig:data_1}, although the FTLE ridges are visible in all reconstructions, the $Lagrangian_{long}$ can support accurate reconstruction of the entire field, whereas the $Lagrangian_{short}$ reconstructions produce minor artifacts in regions of low separation.
%Our trained model performs well for all the testings. 
%
%
%Errors of inference ABC flows are more significant than the Double Gyre as expected since the ABC flow is 3D and more complicated.
%The spatial domain of ABC is much larger than the Double Gyre, and more training samples should be placed to achieve the same effect.          
% In Figure \ref{fig:num_seeds_error_map}, we draw the error map using Double Gyre data sets, which presents our trained model's 
% performance as influenced by the complexity of flow behavior in the domain. Errors are aggregated along the trajectories using AEDR metrics~\cite{ren2020uncertainty}. 
% \mengjiao{compare to FTLE}. Moreover, these error maps confirm that increasing the number of seeds can increase the inference accuracy. 
% In addition, we measure the individual distance between the model-generated results to the ground truth (Figure \ref{fig:num_seeds_violin}). 
% There is a clear trend of decreasing mean and median error by increasing the number of seeds in the domain. 
% Our trained model performs well for all the testings. 
% The models trained with data sets generated from the second approach introduce more errors due to error propagation. 
% Errors of inference ABC flows are more significant than the Double Gyre as expected since the ABC flow is 3D and more complicated.
% The spatial domain of ABC is much larger than the Double Gyre, and more training samples should be placed to achieve the same effect.          

Finally, to assess the inference results qualitatively, Figure \ref{fig:num_seeds_trajectories} shows the model-generated trajectories and 
the ground truth Double Gyre trajectories by varying the number of training seeds. 
The reconstructed results were almost identical to the ground truth for all new trajectories when 10,000 or more seeds were used for training.
When 5,000 seeds were used for training, the $Lagrangian_{short}$ demonstrated lower reconstruction accuracy as interpolation error propagates and accumulates.
In contrast, the $Lagrangian_{long}$ closely followed the ground truth. 
Here, each location along the trajectory was interpolated directly from the starting seed location.
For the $Lagrangian_{long}$, even training data generated using 5,000 seeds were sufficient to maintain accuracy.
% Interestingly, Figure \ref{fig:3b} (d) is observed less accurate trajectories compared to (b) on the left side.
% One possible explanation is the nature of deep learning 
% that every training is not guaranteed the best results. 
%Overall, a more significant number of seeds improve the average accuracy. 

\subsection{Impact of File Cycle Interval}
\label{sec:writing_cycles}

\begin{figure}[]
    \centering
    \subfigure[Resulting error maps when varying the file cycle intervals used to generate training data.]
    {\label{fig:4a}\includegraphics[width=\textwidth]{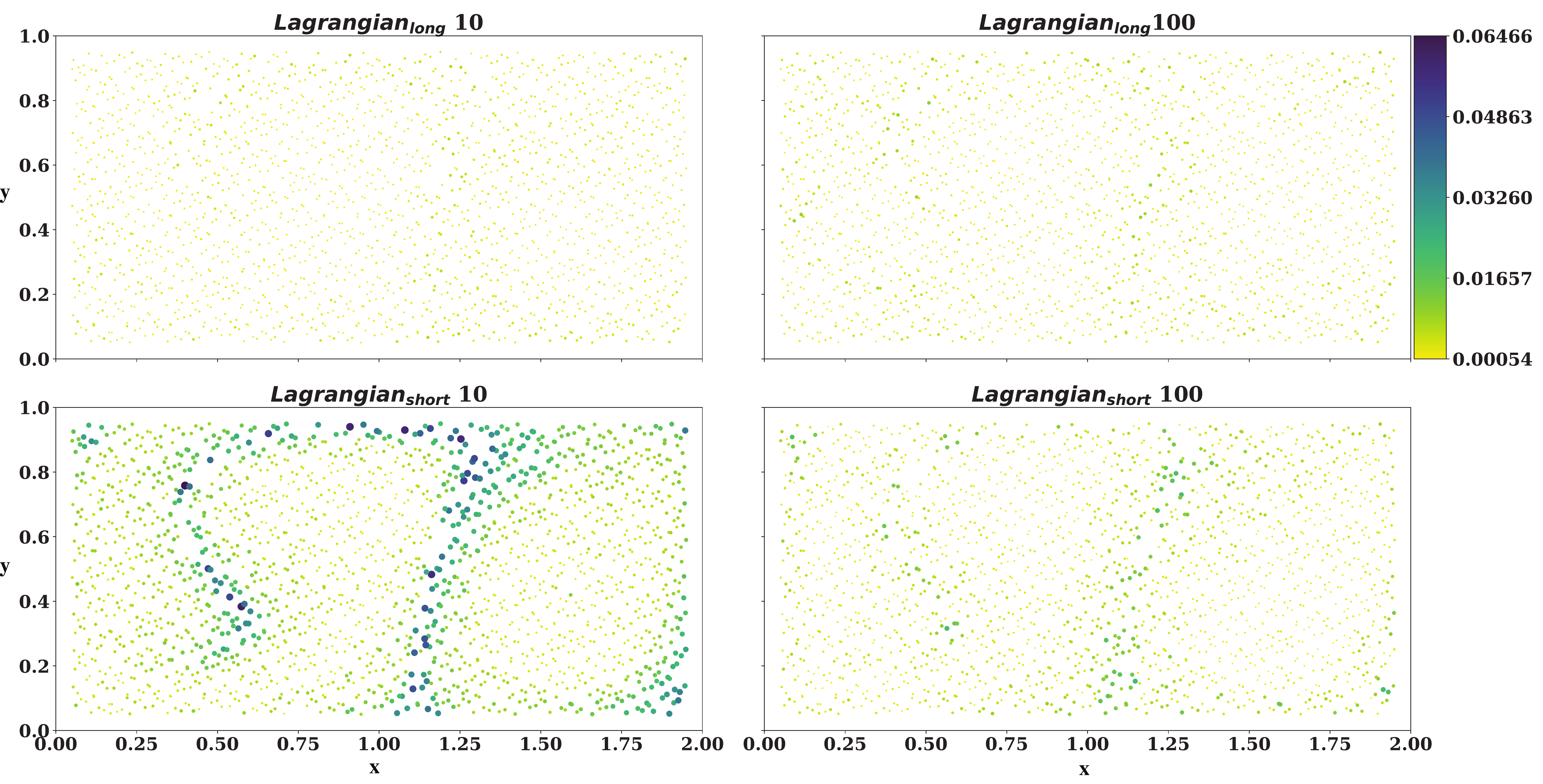}}
    % \quad
    % \subfigure[Double Gyre - Lagrangian\_short - Number of seed 10,000\label{fig:4b}]
    % {\includegraphics[width=\textwidth]{figures/error_maps/short_interval_error_map.pdf}}
    % \subfigure[FTLE computed using inference results of two approaches and the ground truth\label{fig:4c}]
    % {\includegraphics[width=\textwidth]{figures/error_maps/second_ftle.pdf}}
    \subfigure[FTLE scalar field derived using trajectories inferred from the model.]
    {\label{fig:4b}\includegraphics[width=\textwidth]{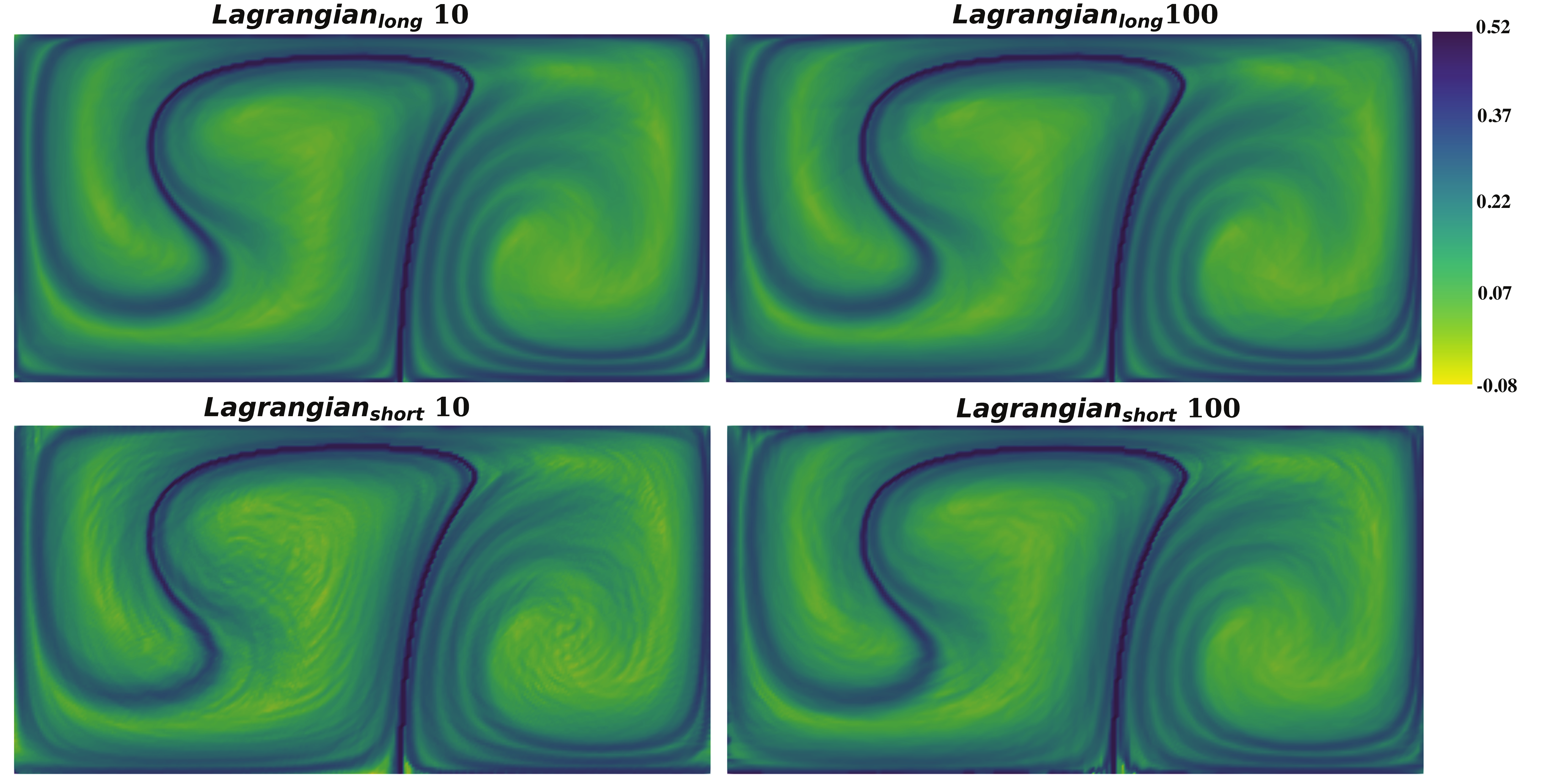}}
%    \subfigure[FTLE of the ground truth]
%    {\label{fig:1c}\includegraphics[width=0.6\textwidth]{figures/ftle.pdf}}
    \caption{\label{fig:writing_cycles_error_map}Visualization of particle trajectory reconstruction error mapped to particle start locations~(\ref{fig:4a}) and the corresponding FTLE scalar fields derived from trajectories inferred by the model~(\ref{fig:4b}), when varying the file cycle interval used to generate training data.
%
%
%The model's performance with varying file cycle intervals is shown as error maps analyzing the model's performance according to the seeds' location in the domain. 
%
Each figure shows the spatial domain $[0, 2] \times [0, 1]$.
The models are trained using 10,000 seeds and the best combination of hyperparameter settings identified in Section~\ref{sec:hyper-parameters}.
We evaluate reconstruction error using 2,000 seeds visualized as circle marks in \ref{fig:4a}.
The color and radius of the circles encode the error aggregated along the trajectories (Equation \ref{eq:measure}).
The top 1\% of errors are treated as outliers and have been removed for analysis from each experiment.
The FTLE is calculated by placing a uniform grid with size $[256 \times 128]$.
The model's performance is related to the flow behavior in the domain, and reconstruction errors are higher in regions with greater separation.
Notably, $Lagrangian_{short}$ tests with a short interval suffer from error propagation and accumulation.}
\end{figure}

To understand the performance of our model with varying file cycle intervals, we evaluated four intervals, 10, 20, 50, and 100,
in our experiments. 
We considered a total of 1,000 cycles of the Double Gyre data set. 
Further, we used a fixed number of 10,000 seeds to generate the training data sets. 

\begin{figure}[]
    \label{fig:5a}
    \includegraphics[width=\textwidth]{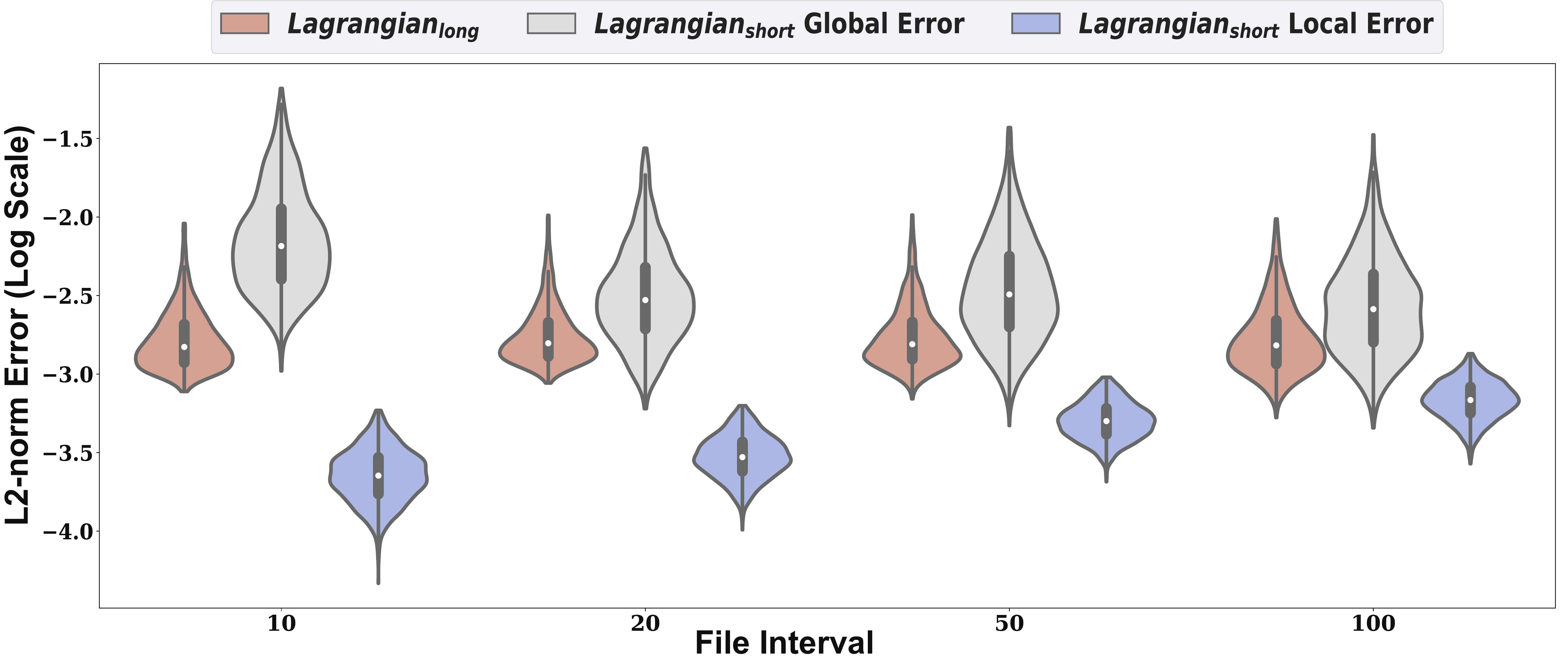}
    % \quad 
    % \subfigure[Double Gyre - Lagrangian\_short - Number of seeds 20,000.\label{fig:5b}]
    % {\includegraphics[width=0.48\textwidth]{figures/interval_violin_plot.pdf}}
    \caption{\label{fig:writing_cycles_violin}Violin plots of inference error evaluated for models trained using data generated for varying file cycle intervals. 
    The errors are calculated along the trajectories using Equation \ref{eq:measure}. The labels on the y-axis use $10^X$ format to show the error. The error is shown as a distribution using violin plots with the minimum, maximum, and median errors. The evaluation is performed using 2,000 random test seeds. 
    The top 1\% of errors are treated as outliers and have been removed for analysis from each experiment.
Although the accuracy of $Lagrangian_{long}$ does not varying significantly with the considered file cycle intervals for the Double Gyre, the global error of the model trained using $Lagrangian_{short}$ decreases in accuracy as the length of the file cycle interval increases, but the local error increases with longer integration durations between file cycles.}
\end{figure}

\begin{figure}[]
    \centering
    \label{fig:9a}
    \includegraphics[width=\textwidth]{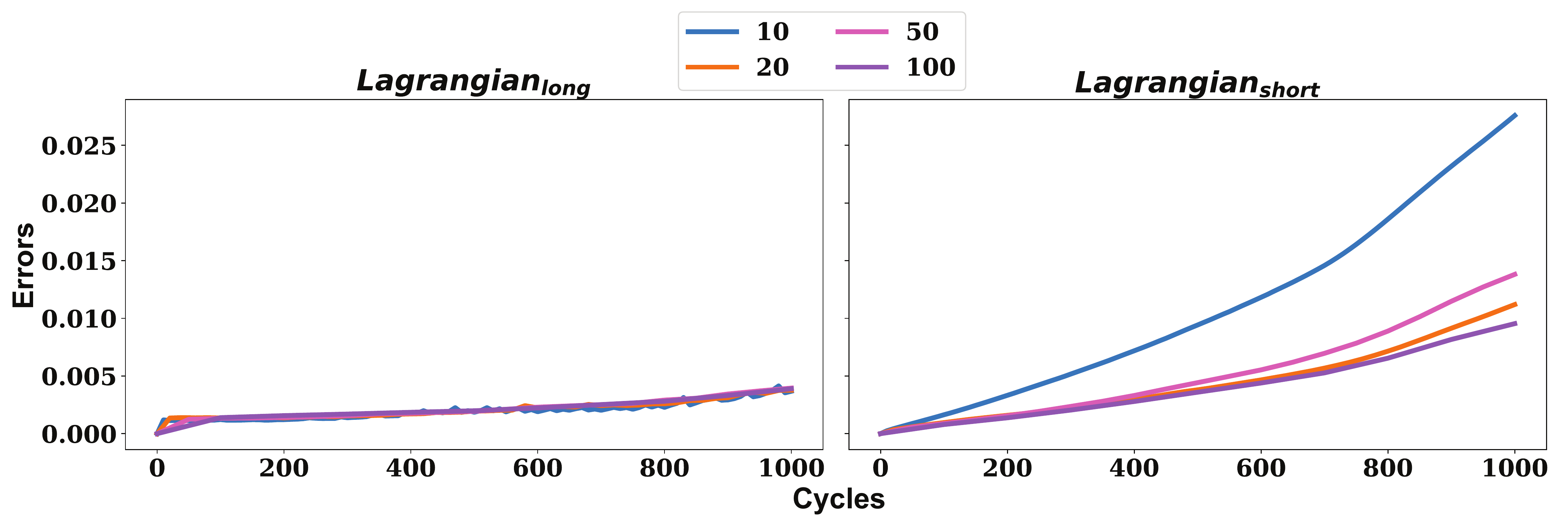}
    \caption{\label{fig:error_ts}The average reconstruction error over file cycles for the Double Gyre data set with varying file cycle intervals. 
    The errors are calculated by averaging distances between the model generated end locations and the ground truth at each file cycle. Evaluations are performed over 2,000 test seeds. 
    For the $Lagrangian_{long}$ approach, errors do not propagate over file cycles. 
    Results of different file cycle intervals have a similar trend.
    In contrast, errors are propagated in the $Lagrangian_{short}$ approach, and shorter file cycle interval results in more significant errors over time.}
\end{figure}

In Figure~\ref{fig:writing_cycles_error_map}, we report the error maps as well as the FTLE derived from using various configurations for training data generation.
The $Lagrangian_{long}$ was not impacted by the file cycle interval since each interpolation was independent of prior locations stored along the trajectory.
Reconstruction of new trajectories using the model trained by the $Lagrangian_{short}$ data involved an interpolation process where each location along the trajectory was dependent on the previous location.
Thus, we observed a higher reconstruction error when the interval was short, and more intervals need to be spanned to construct a trajectory over the entire temporal duration.
For example, for training data generated by the $Lagrangian_{short}$ using an interval of 10, we saw the reconstruction error was higher for particles originating near FTLE ridges.
These findings are consistent with the error analysis of Lagrangian-based particle tracing systems (\cite{chandler2016analysis}).
Similar to prior experiments, in Figure~\ref{fig:data_1}, we observed the derived FTLE scalar fields are accurate for the $Lagrangian_{long}$, but contained some artifacts for the $Lagrangian_{short}$.
Here, as expected, the $Lagrangian_{short}$ shows fewer artifacts when using a longer file cycle interval.

Considering the violin plots in Figure~\ref{fig:writing_cycles_violin}, we obsersed varying reconstruction accuracy patterns.
The $Lagrangian_{long}$ accuracy did not change significantly with the file cycle interval.
The local error of the $Lagrangian_{short}$ was low for short intervals, but increased as the interval length increased due to greater divergence between neighboring trajectories over longer integration times.
The global error of the $Lagrangian_{short}$ represented the accuracy of particle trajectories that are ``stitched''.
We found the global error was the highest when the file cycle interval was short given a greater number of ``stitching'' events were involved. 
As the file cycle interval increased, although the accuracy of every individual interpolation (local error) was higher, the global error decreased due to fewer total interpolation steps.
Again, these findings are consistent with prior work by \cite{chandler2016analysis} and \cite{sane2019interpolation}.
Additionally, we present the average error across all particles over time for the $Lagrangian_{long}$ and $Lagrangian_{short}$ approaches in Figure~\ref{fig:error_ts}.
The line curves provide strong evidence of local error propagation and accumulation for tests using $Lagrangian_{short}$ training data.

\begin{figure}[!htb]
    \label{fig:7a}
    \includegraphics[width=\textwidth]{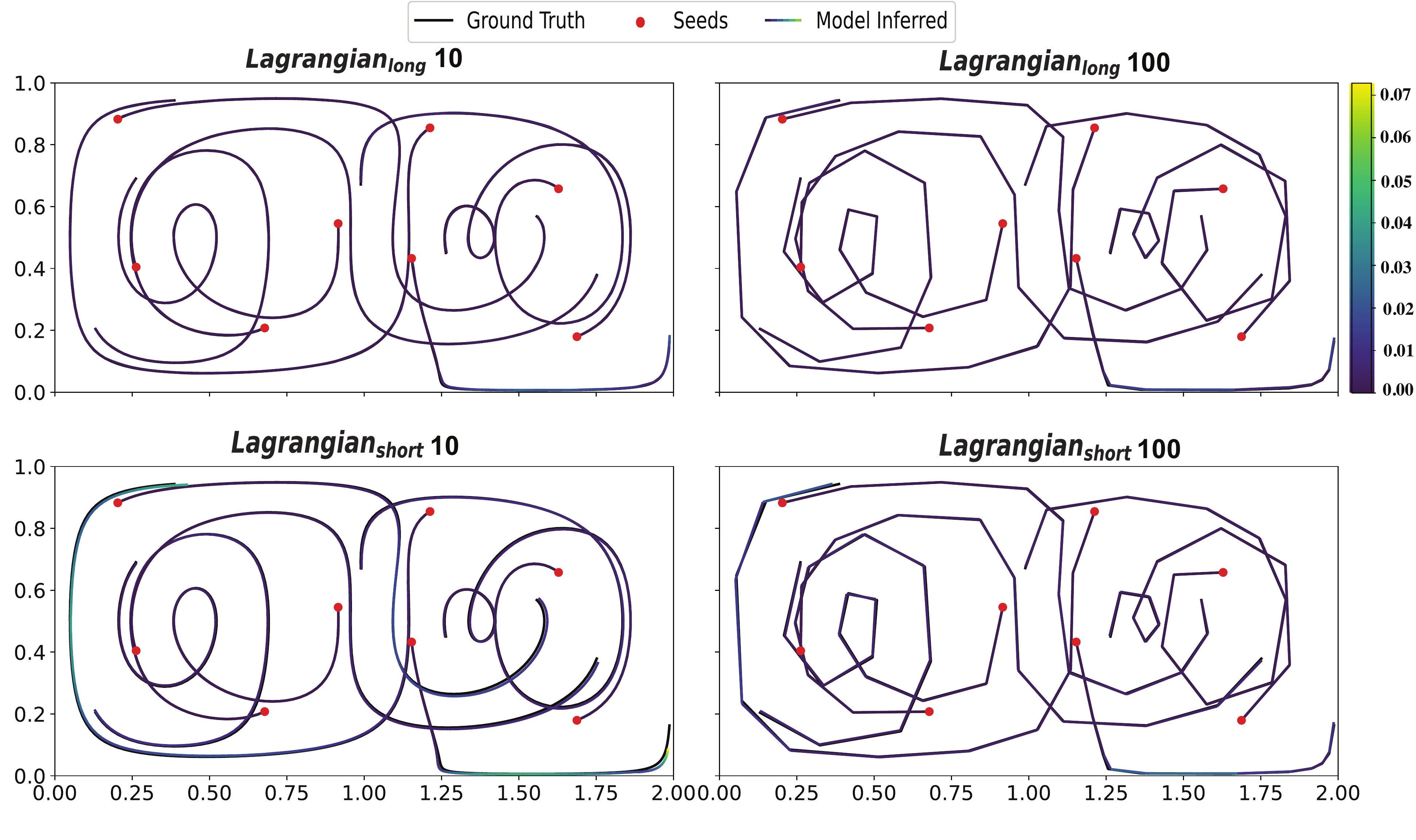}
    % \quad
    % \subfigure[Double Gyre - Lagrangian\_short - Number of seeds 20,000.\label{fig:7b}]
    % {\includegraphics[width=\textwidth]{figures/multiple_fm_seeds_trajectories_interval.pdf}}
    \caption{\label{fig:writing_cycles_trajectories}Visualization of inferred trajectories and the ground truth for the Double Gyre with different file cycle intervals. 
    The seeds were randomly placed using the Sobol seeding strategy. 
    The colors of model inferred trajectories indicate the distance between the model inferred end location and the ground truth. 
    Our trained model can reconstruct trajectories almost visually identical to the ground truth.}
\end{figure}

For a qualitative assessment of the impact of the file cycle interval, we present reconstructed pathlines alongside the ground truth in Figure~\ref{fig:writing_cycles_trajectories}.
We used piecewise linear interpolation to connect every interpolated location along the new trajectories.
Although the $Lagrangian_{short}$ demonstrated a small deviation from the ground truth when short file cycle intervals were used, the overall accuracy of reconstructed trajectories was high with interpolated results closely overlapping the ground truth.
%
%The visualization result of the model-generated trajectories and the ground truth 
%with varying file cycle interval is shown in Figure \ref{fig:writing_cycles_trajectories}. 
%The reconstructed results are almost identical to the ground truth for all shown cases.  

\subsection{Application to Fluid Dynamics Machine Learning Data Set}

\begin{figure}[]
    % \subfigure[Resulting error maps of Gerris flow.\label{fig:gerris-trajectories}]
    \includegraphics[width=\textwidth]{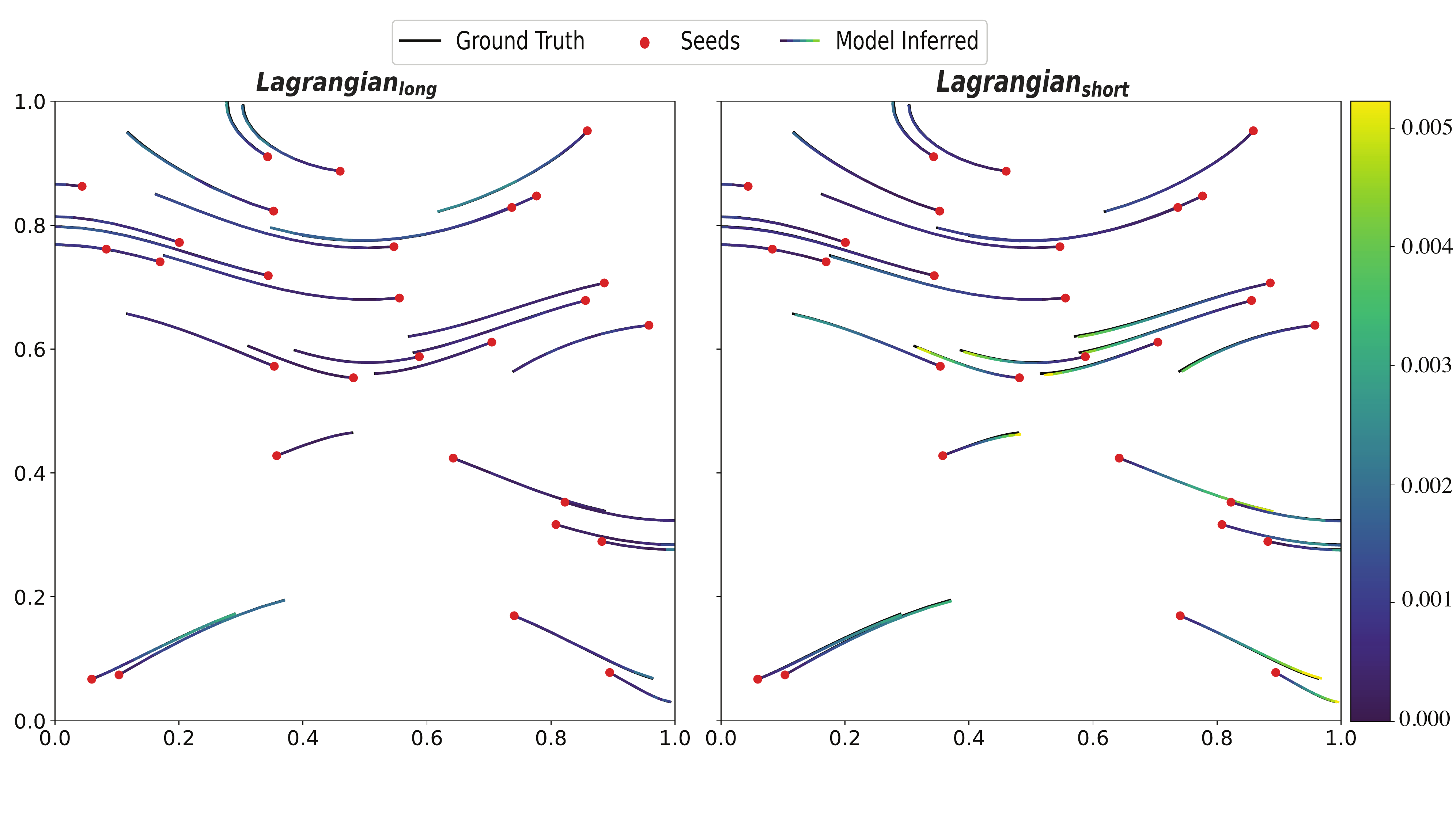}
    \caption{\label{fig:gerris-trajectories}Visualization of inferred trajectories and the ground truth for the ensemble member \#200 vector field.}
\end{figure}

We applied our method to an ensemble member (\#200) of the 2D fluid dynamics machine learning data set generated using the Gerris flow solver (\cite{Jakob2020}).
The resolution of the original data set is $[512 \times 512 \times 1001]$.
To generate the training data set, we placed $50,000$ seeds in the domain, set the file cycle interval to 10, and traced flow maps over the first 100 cycles. 
For particle advection, we used the VTK-m (\cite{moreland2016vtk}) library and a fourth-order Runge-Kutta (RK4) advection kernal. 
%
% In Figure \ref{fig:gerris}, we report the spatial error maps and the violin plots of the reconstruction error when using training data set generated by the $Lagrangian_{long}$ and $Lagrangian_{short}$ approaches. 
%
The median error of using our method after 100 cycles and 10 interpolation steps is approximately two times the grid cell size. 
Our method cost 0.6 seconds for reconstructing 2,000 particle trajectories using parallel inferences with OpenMP (\cite{dagum1998openmp}).
%
%However, using VTK-m costs 99 seconds for fetching velocity fields and finishing particle tracing parallel with OpenMP (\cite{dagum1998openmp}). 
%
When considering the storage requirements, the subset of the original data size we consider is approximately 209MB. 
Since our model has a fixed memory requirement, once trained, the storage costs are still fixed at 10.5 MB. 
To qualitatively evaluate the reconstructed data, we visualize pathlines inferred by the trained model in comparison with the ground truth in Figure~\ref{fig:gerris-trajectories}. 
%
%Since the ensemble member we consider has significant turbulence, we observe reconstruction accuracy reduces along the trajectory for both approaches.
%
In future works, we aim to study how to improve inteprolation accuracy as well as determine an appropriate number of samples to be computed using in situ processing.

% \begin{figure}[!htb]
%     \centering
%     \subfigure[Resulting error maps of Gerris flow.\label{fig:gerris-errormap}]
%     {\includegraphics[width=\textwidth]{figures/gerris/gerris_error_maps.pdf}}
%     \subfigure[Inference error distribution of Gerris flow.\label{fig:gerris-violin}]
%     {\includegraphics[width=0.8\textwidth]{figures/gerris/gerris_violin_plot.pdf}}
%     \caption{\label{fig:gerris}Visualization of particle trajectory reconstruction error mapped to particle start locations (\ref{fig-errormap}) and 
%     inference errors distribution evaluated by the trained model (\ref{fig:gerris-violin}). 
%     Errors are calculated along the trajectories using AEDR metrics and shown as circles' color and radius in \ref{fig-errormap} and 
%     violin plots with minimum, maximum, and median errors in \ref{fig:gerris-violin} and evaluated over 2000 random seeds. 
%     The top 1\% of errors are treated as outliers and havebeen removed foranalysis from each experiment. Our method generate median error of 0.004, which about 2X cell size.}
% \end{figure}

\section{Future Work and Conclusion}
Exploratory flow visualization for large-scale time-varying vector field data is challenging.
In this paper, we introduced a deep neural network-based approach using Lagrangian represesntations to enable exploratory analysis.
Our study demonstrated our model can be trained using Lagrangian representations extracted from a 2D
time-varying vector field.
Specifically, we used the widely studied unsteady Double Gyre analytical flow data set and one fluid dynamics machine learning data set to demonstrate our method.
We contributed the first assessment of applying deep learning to various forms of Lagrangian representations and evaluated the efficacy of exploratory analysis.
%
%This allows us to reason more effectively about approaches to reduce reconstruction interpolation error when using machine learning.
%
A benefit of using our method is the fixed memory required by a model and fast inference of unstructured spatiotemporal data.
Our trained model requires only 10.5 MB, and consequently, time spent on I/O to load the model during post hoc analysis is negligible.
Further, we are able to infer the pathlines of thousands of particles at interactive rates.
With respect to reconstruction interpolation error, we found inference errors are small and follow predictable patterns consistent with results from prior works. 
Predictable and consistent error patterns enable effective future navigation of strategies to reduce reconstruction interpolation error when using machine learning.
Overall, our study demonstrates the benefits of leveraging deep learning for exploratory flow visualization of time-varying vector field data.

An important direction for future work is investigating model performance for more complex or turbulent flows as well as large-scale 3D flow fields.
%
%A potential solution to address limitations could be to create a distributed system of models for spatiotemporal blocks of data.
%
With the objectives of improving spatial and temporal interpolation accuracy 
and reducing model training time, various forms of training data to train a model or different network architectures could be considered.
For example, concatenate sets of $Lagrangian_{long}$ trajectories to limit instances of error propagation while simultaneously accounting for reduced interpolation error due to stretching or divergence in the flow. 
Lastly, an open-source interactive tool for interactive flow visualization exploration, with a trained
model serving as a backend, would be valuable to the community. 
We plan to pursue these projects
in the future.

\acknowledgements
The authors acknowledge current research support provided in part by the Intel Graphics and Visualization Institutes of XeLLENCE, 
the National Institutes of Health under grant numbers P41 GM103545 and R24 GM136986, 
the Department of Energy under grant number DE- FE0031880, and the Utah Office of Energy Development.

%% appendix sections are then done as normal sections and after Acknowledgements



% \begin{thebibliography}{-8}

% % \bibitem must have the following form:

% % \small{
% % \bibitem{key}

% % ...

% % }

% \end{thebibliography}


\bibliographystyle{Bibliography_Style}
% Bibliography_Style

\bibliography{mhan_jfvip21}
\end{document}